\pdfoutput=1
\documentclass[10pt,logo,copyright]{nvidiatechreport}
\usepackage[numbers,sort&compress,square]{natbib}

\usepackage[utf8]{inputenc} %
\usepackage[T1]{fontenc}    %

\usepackage{parskip}        %
\usepackage{url}            %
\usepackage{hyperref}       %
\usepackage{booktabs}       %
\usepackage{amsfonts}       %
\usepackage{nicefrac}       %
\usepackage{microtype}      %
\usepackage{xcolor}         %
\usepackage[dvipsnames]{xcolor} %
\usepackage{graphicx}
\usepackage{animate}        %
\usepackage{subcaption}
\usepackage{tabularx}
\usepackage{makecell}
\usepackage{adjustbox}
\usepackage{setspace}
\newcolumntype{M}[1]{>{\centering\arraybackslash}m{#1}}
\usepackage{float}
\usepackage{tikz}
\usetikzlibrary{positioning,shapes,arrows}
\usepackage{amsmath,amsfonts,bm, bbm,leftindex}
\usepackage{multirow}
\usepackage{comment}
\usepackage{gensymb}
\usepackage{lipsum}
\usetikzlibrary{arrows.meta, positioning, fit}
\usepackage[para]{threeparttable}
\usepackage{tikz}
\usetikzlibrary{tikzmark}

\usepackage{placeins}

\usepackage{pifont}
\usepackage{wrapfig}
\usepackage{caption}
\usepackage{algorithm}
\usepackage{algpseudocode}
\usepackage{amsmath}
\usepackage{amssymb}
\usepackage{graphicx}
\usepackage{xcolor}
\newcommand{\name}{Flora\xspace}
\newcommand{\hflogo}{\raisebox{-1.5pt}{\includegraphics[height=1.05em]{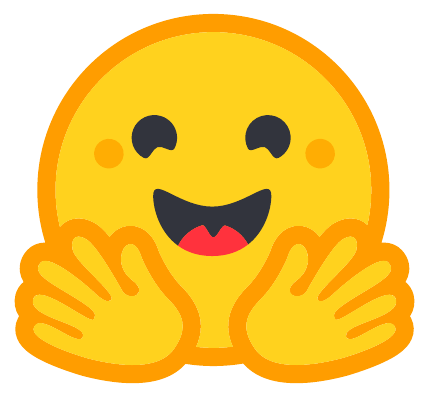}}}

\input{assets/fonts/t1FloraNV.fd}
\DeclareFontFamily{TS1}{FloraNV}{}
\DeclareFontShape{TS1}{FloraNV}{m}{n}{<-> ssub * lmr/m/n}{}
\pdfmapline{+FloraNV-Rg <NVIDIA-Sans-Font-TTF/NVIDIASans_Rg.ttf <assets/fonts/FloraT1.enc}
\pdfmapline{+FloraNV-Bd <NVIDIA-Sans-Font-TTF/NVIDIASans_Bd.ttf <assets/fonts/FloraT1.enc}
\pdfmapline{+FloraNV-It <NVIDIA-Sans-Font-TTF/NVIDIASans_It.ttf <assets/fonts/FloraT1.enc}
\pdfmapline{+FloraNV-BdIt <NVIDIA-Sans-Font-TTF/NVIDIASans_BdIt.ttf <assets/fonts/FloraT1.enc}
\newcommand{\floraChartFont}{\fontfamily{FloraNV}\selectfont}

\definecolor{floraFirst}{HTML}{E5E9E3}
\definecolor{floraOurFirst}{HTML}{DCE9CE}
\definecolor{floraGreen}{HTML}{466C27}
\newcommand{\floraRankFirst}[1]{\cellcolor{floraFirst}\textbf{#1}}
\newcommand{\floraOurScore}[1]{\textcolor{floraGreen}{#1}}
\newcommand{\floraOurBest}[1]{\cellcolor{floraOurFirst}\textcolor{floraGreen}{\textbf{#1}}}

\newcommand{\floraTableStyle}{%
\linespread{1}\small\floraChartFont\setlength{\tabcolsep}{4pt}\renewcommand{\arraystretch}{1.05}%
\captionsetup{justification=raggedright,singlelinecheck=false,font={small,rm},skip=5pt}}

\newcommand{\rewardPending}{\textcolor{black!45}{\textemdash}}
\newcommand{\rewardTableStyle}{%
\linespread{1}\small\floraChartFont\setlength{\tabcolsep}{4pt}\renewcommand{\arraystretch}{1.06}%
\captionsetup{justification=raggedright,singlelinecheck=false,font={small,rm},skip=5pt}}

\usepackage[most]{tcolorbox}
\usepackage{xcolor}
\usepackage{hyperref}
\definecolor{boxaccent}{HTML}{76B900}
\tcbset{
linkbox/.style={
colback=black!2,
colframe=boxaccent!75!white,
boxrule=0.4pt,
arc=1pt,
left=4pt, right=4pt, top=4pt, bottom=4pt,
}
}

\usepackage[nameinlink]{cleveref}
\crefname{equation}{Eq.}{Eqs.}
\crefname{figure}{Fig.}{Figs.}
\crefname{section}{Sec.}{Sec.}
\crefname{appendix}{App.}{App.}
\crefname{table}{Tab.}{Tabs.}
\crefname{algorithm}{Algo}{Algo}

\title{FLORA: An Open Model Suite for\\Scaling Robotics Data Engines}

\author{\makebox[\linewidth][l]{\parbox[t]{\dimexpr\linewidth+24pt\relax}{\raggedright Sicong Leng$^{*}$, Yunze Man$^{*}$, Zhiqi Li$^{*}$, Jing Wang, Yuhe Zhang, Ryan Punamiya, Siddharth Patki, Kaiyuan Zheng, Zi-ang Cao, Zhiding Yu, Sifei Liu, Danfei Xu, Ben Levin, Jan Kautz, Yuke Zhu$^{\dagger}$, Johan Bjorck$^{*\dagger}$, Linxi ``Jim'' Fan$^{\dagger}$}}}

\renewcommand{\titlefont}{\color{nvidiagreen}\normalfont\rmfamily\bfseries\fontsize{24}{27}\selectfont}
\hypersetup{allcolors=floraGreen}
\renewcommand{\absfont}{\normalfont\linespread{1.12}\fontsize{10}{12}\selectfont}

\begin{document}
\renewcommand{\rmdefault}{XCharter-TLF}
\renewcommand{\sfdefault}{FloraNV}
\renewcommand{\familydefault}{\rmdefault}
\normalfont
\fancyhead[C]{\footerfont FLORA: An Open Model Suite for Scaling Robotics Data Engines}
\hypersetup{pdftitle={FLORA: An Open Model Suite for Scaling Robotics Data Engines}}
\maketitle
\vspace{-5mm}
{\linespread{1}\fontsize{9}{11}\selectfont\sffamily\urlstyle{same}
\setlength{\tabcolsep}{0pt}\renewcommand{\arraystretch}{1}
\begin{tabularx}{\linewidth}{@{}X@{\hspace{18pt}}X@{}}
\href{https://research.nvidia.com/labs/gear/flora/}{\textcolor{black!80}{\textbf{Blog Post}}} &
\href{https://huggingface.co/collections/nvidia/flora}{\textcolor{black!80}{\hflogo\enspace\textbf{Open Model Suite}}} \\
\href{https://research.nvidia.com/labs/gear/flora/}{\nolinkurl{research.nvidia.com/labs/gear/flora/}} &
\href{https://huggingface.co/collections/nvidia/flora}{\nolinkurl{huggingface.co/collections/nvidia/flora}}
\end{tabularx}\par}
\vspace{-3mm}

\begin{abstract}
The life-cycle of robotics foundation models crucially depends on data labels. Outcome reward labels are needed for monitoring real-world deployments. Process reward labels are needed during RL post-training. Timestamped captions are needed for filtering and balancing pretraining data. We present \name, an industrial-quality data engine used for training GR00T models, that simultaneously covers these capabilities. \name achieves SOTA scores on the captioning benchmark WGO-Bench and reward benchmarks RoboReward and Robometer. On internal benchmarks it is competitive with GPT-6 Astra while beating other frontier models. The model is substantially cheaper than API-based alternatives, roughly $\sim125\times$ cheaper than GPT-6 Astra. \name is battle tested at scale, labeling 200,000 hours of data per day across 512 GB300s.  By open-sourcing our data engine, we enable researchers to scale custom robotics datasets and explore emerging applications such as physical auto-research and human-to-robot transfer.

\end{abstract}
\abscontent
\begin{figure}[H]
  \centering
  \vspace{-2mm}
  \includegraphics[width=\linewidth]{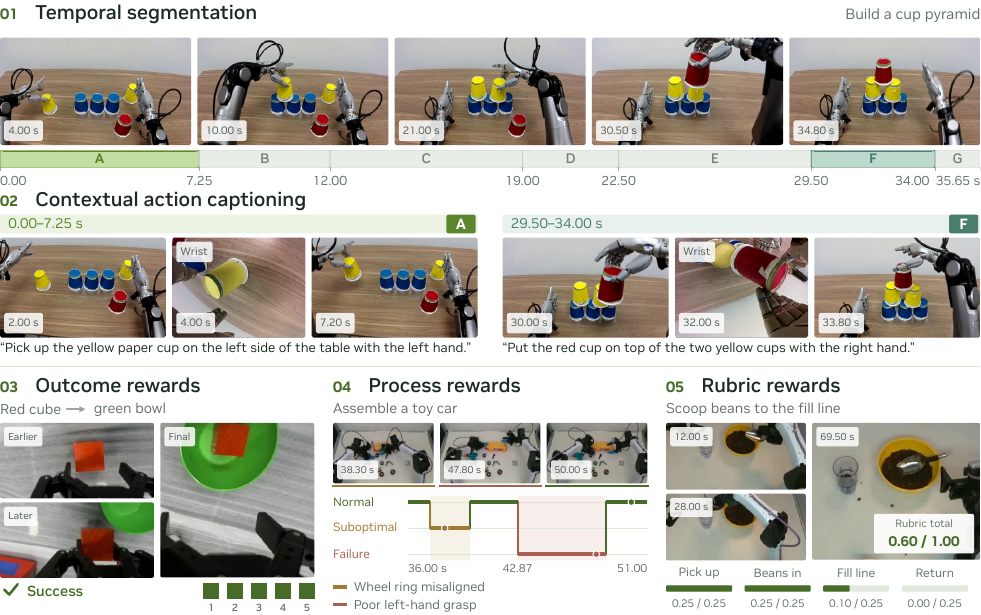}
  \vspace{-5mm}
  \caption{\textbf{Capabilities.} \name produces timestamped captions by first segmenting videos and then captioning individual segments. \name can also produce  outcome and process rewards, and rewards based on rubrics specified at test-time.}
  \label{fig:release-gallery}
\end{figure}
\clearpage

\begin{figure}[H]
  \centering
  \captionsetup{skip=9pt}
  \includegraphics[width=\linewidth]{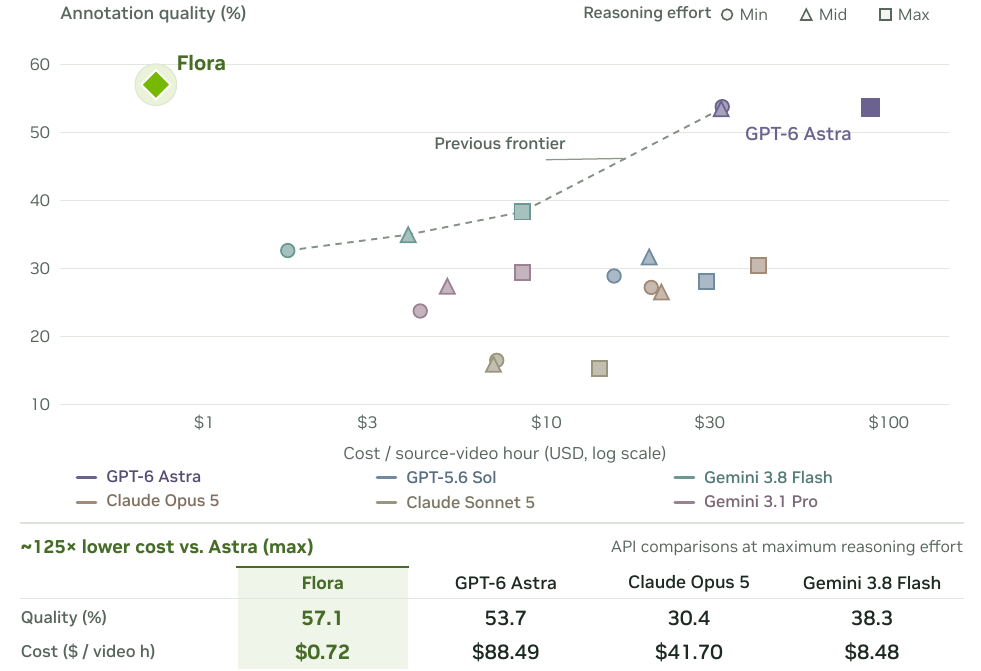}
  \caption{\textbf{Annotation quality versus cost.} \name achieves competitive quality at substantially lower cost compared to GPT-6 Astra. Evaluation and cost assumptions appear in Appendix~\ref{sec:annotation-cost-estimation}.}
  \label{fig:annotation-cost}
  \label{fig:annotation-efficiency}
\end{figure}

\section{Overview}
\label{sec:intro}

Robotics foundation models have evolved from single-embodiment policies~\citep{brohan2022rt1} to generalist models trained on large, heterogeneous corpora~\citep{zitkovich2023rt2,intelligence2025pi05,bjorck2025gr00t,geminirobotics2025}. Across these large datasets, high-quality labels are imperative for performant models. \emph{Pretraining} benefits from timestamped action instructions that enable language conditioning~\citep{xiao2022dial} and let practitioners rebalance the data mixture~\citep{hejna2024remix,xie2023doremi}. \emph{RL post-training}~\citep{li2025simplevlarl,intelligence2025pistar06} requires \emph{process rewards} that measure progress within an episode, analogous to step-level supervision in language models~\citep{uesato2022solving,lightman2023verify}. \emph{Deployment} needs \emph{outcome rewards} to monitor fleets, detect failures~\citep{duan2024aha}, and evaluate policies at scale~\citep{atreya2025roboarena}.

Most existing systems cover only one annotation type. General-purpose frontier models can provide several types, but are expensive at the scale of modern robot datasets. We present \name, an industrial-quality data engine used to train GR00T models. It consists of three models---a captioning model, a segmentation model, and a reward model---that together produce timestamped captions, process rewards, and outcome rewards. User-defined rubrics additionally support flexible reward modeling at test time. The engine is battle tested at scale, labeling up to 200,000 hours of data per day across 512 GB300 GPUs. By open-sourcing \name, we enable researchers to scale custom robot datasets and explore emerging applications such as physical auto-research~\citep{lu2024aiscientist} and human-to-robot transfer~\citep{kareer2024egomimic}. 

\clearpage

\section{Capabilities and Results}
\label{sec:eval}

\begin{table}[H]
  \centering
  \caption{\textbf{Benchmark scores.} \name is competitive with GPT-6 Astra while beating other baselines. Values are scaled by 100. Full comparisons appear in Appendix~\ref{sec:complete-results}.}
  \label{tab:release-scorecard}
  \includegraphics[width=\linewidth]{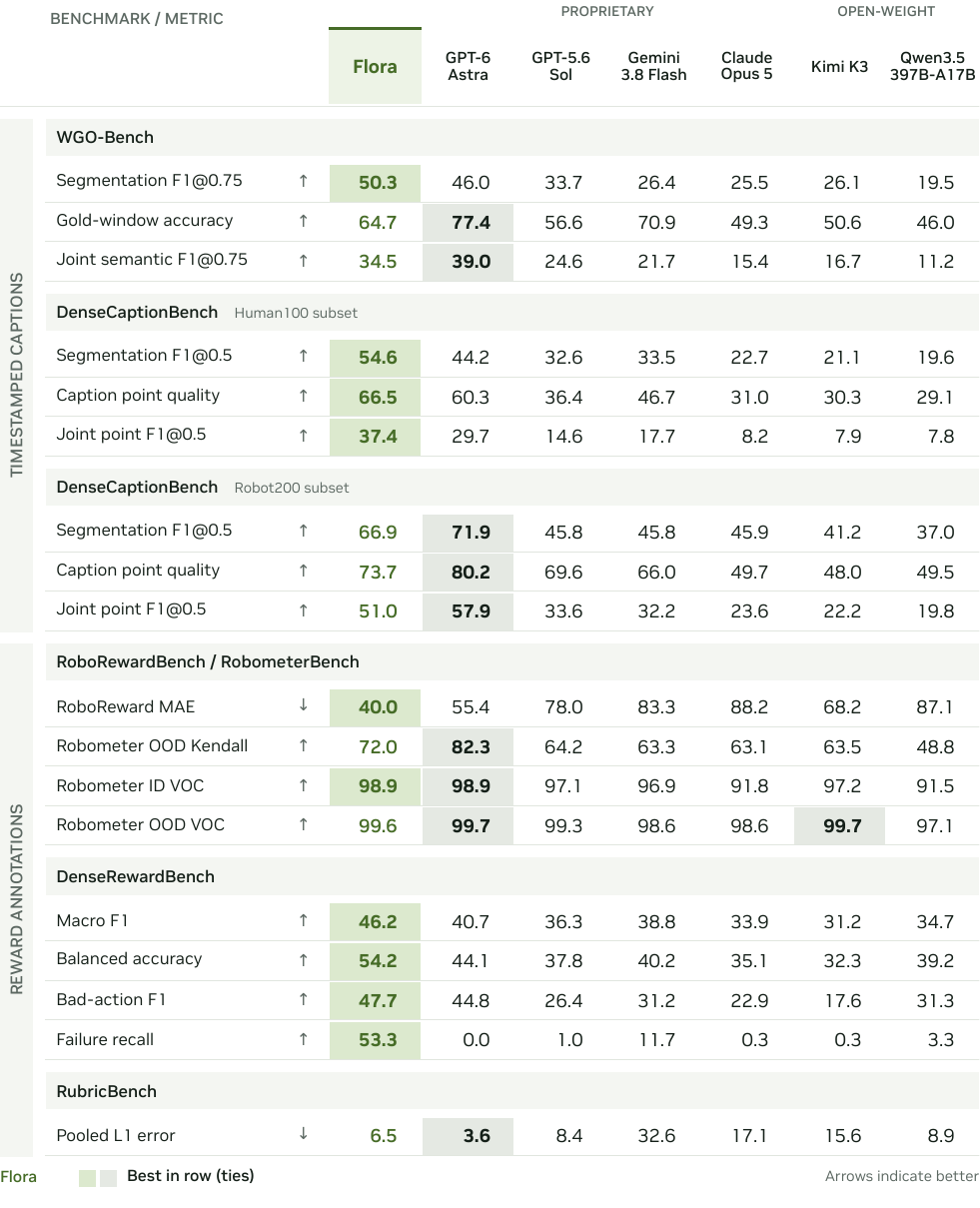}
\end{table}

\begin{figure}[H]
  \centering
  \includegraphics[width=\linewidth]{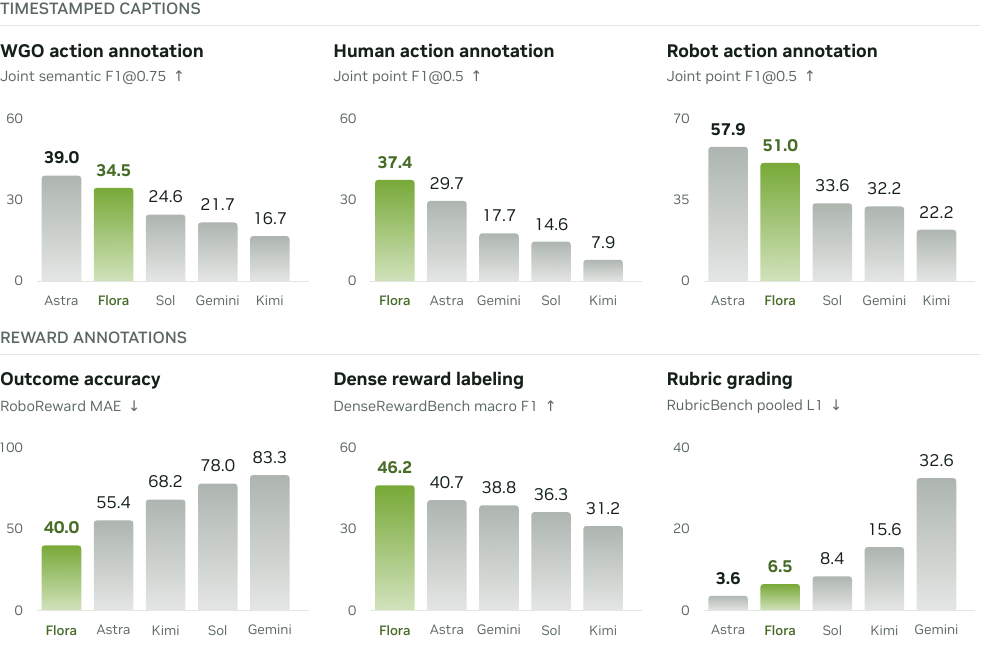}
  \caption{\textbf{One data engine, complementary capabilities.} Timestamped caption and reward labeling results across six tasks. All values are scaled by 100. Full comparisons appear in Appendix~\ref{sec:complete-results}.}
  \label{fig:release-capabilities}
\end{figure}

\subsection{Timestamped Action Captions}
\label{sec:textual-annotation-eval}

We evaluate \textit{temporal segmentation}, \textit{captioning} on gold reference intervals, and \textit{joint annotation} on predicted intervals. \name uses its default pipeline (Section~\ref{sec:textual-annotation}) with the validation-selected segmentation and captioning RL checkpoints.

\phantomsection\label{par:common-pipeline}
\textbf{Common pipeline.} We use a shared two-stage pipeline for off-the-shelf models. One call segments timestamped episode contact sheets, followed by one captioning call per fixed interval with up to two seconds of context on either side. Both stages use the same model and fixed stage-specific prompts. Caption-only evaluation substitutes reference intervals. Reference captions are never supplied. These comparisons evaluate complete systems with different visual inputs and inference compute (Appendix~\ref{sec:annotation-protocol}).

\textbf{WGO-Bench.} Table~\ref{tab:wgo} reports model comparisons on WGO-Bench~\citep{macrodata2026wgo}, comprising 25 human egocentric episodes (HomER), 50 robot episodes with external views (DROID), and 25 with head-camera views (Galaxea). All systems, including \name, receive the episode instruction in both stages. Comparisons use one-to-one temporal matching at tIoU $\geq0.75$ and the public text-only caption rubric. Joint matches require both temporal and caption agreement. \name attains 50.3\% mean segmentation F1, 34.5\% mean joint F1, and 64.7\% gold-window caption accuracy. It leads segmentation and ranks second in joint F1 behind Astra's 39.0\%. Caption accuracy remains below Astra and Gemini 3.8 Flash, so the gains primarily reflect temporal localization.

\begin{figure}[!htbp]
  \centering
  \includegraphics[width=\linewidth]{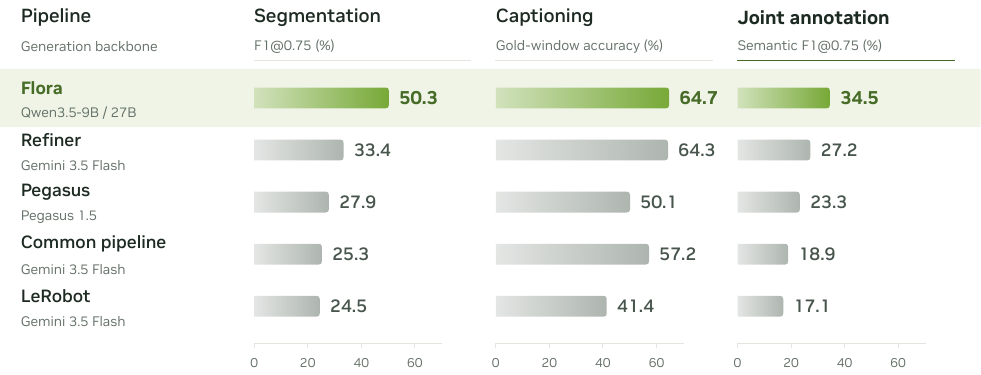}
  \caption{\textbf{Complete pipelines on WGO-Bench.} Equal-subset mean scores (\%, $\uparrow$), with generation backbones shown. Full results and protocols appear in Appendix~\ref{sec:wgo-pipelines}.}
  \label{fig:wgo-pipelines}
\end{figure}

\textbf{Pipeline comparison.} Figure~\ref{fig:wgo-pipelines} compares \name with the common pipeline, Refiner~\citep{macrodata2026refiner}, LeRobot~\citep{lerobot2026v060}, and Pegasus~\citep{twelvelabs2026pegasus15}. Compared with Refiner, \name improves mean segmentation F1 by 16.9 points and joint F1 by 7.3 points, with similar gold-window caption accuracy (64.7\% vs.\ 64.3\%). Additional backbone comparisons and API substitution diagnostics appear in Appendix~\ref{sec:wgo-pipelines}.

\phantomsection\label{sec:internal_cap_benchmark}
\textbf{DenseCaptionBench.} WGO-Bench provides a public test of subtask annotation, but its 100 single-view episodes offer limited task and scenario coverage and do not test multi-view recordings or dexterous-hand manipulation. Its binary caption matches also do not quantify partial coverage of action details. We introduce DenseCaptionBench to assess these requirements on 300 complete episodes, with 100 each for humans, parallel-gripper robots, and dexterous-hand robots. Robot episodes retain all camera views. Selection prioritizes task, action, and manipulator-role coverage without using model performance scores. The human track spans 56 environment labels and 95 task labels. Segmentation reports F1 at tIoU $\geq0.5$. Oracle captioning gives partial credit for intended action, object identity, manipulator coordination, and spatial relations while penalizing demonstrated factual errors. Appendix~\ref{sec:annotation-protocol} details construction and scoring.

\Needspace{3\baselineskip}
Table~\ref{tab:internal-e2e} reports Human100 and Robot200 results. \name obtains 54.6\% segmentation F1, 66.5\% oracle caption quality, and 37.4\% joint F1 on Human100, leading all three metrics. Robot Avg scores are 66.9\%, 73.7\%, and 51.0\%, respectively. Human joint F1 exceeds Astra's 29.7\% by 7.7 points. On Robot200, \name ranks second to Astra across the three metrics. Its 51.0\% joint F1 compares with 57.9\% for Astra and 33.6\% for the next-best baseline.

\subsection{Reward Labeling}
\label{sec:reward-eval}

We evaluate \name-Rewarder on outcome, dense, and rubric-derived rewards across two public and two internal benchmarks. General-purpose baselines receive the same task instructions and answer formats as \name. Frame budgets and decoding settings are model-specific. Results for specialized reward models are taken from their publications and retain their original protocols. Further details are given in Appendix~\ref{sec:reward-eval-protocol}.

\Needspace{3\baselineskip}
\textbf{Outcome accuracy.} RoboRewardBench~\citep{lee2026roboreward} scores 2,831 episodes on a human-labeled 1--5 outcome scale, and we report the mean absolute error (MAE) of the decoded score. As Table~\ref{tab:public-reward} shows, \name-Rewarder achieves an MAE of 40.0, compared with 55.4 for GPT-6 Astra and 66.5 for the specialized RoboReward model, with all errors multiplied by 100. This is a 27.8\% reduction relative to Astra. The advantage comes from partially completed tasks.

\Needspace{3\baselineskip}
\textbf{Trajectory ranking and progress.} RobometerBench~\citep{robometer2026} tests whether rewards rank 567 out-of-distribution (OOD) trajectories by quality, measured by Kendall's $\tau_a$, and whether predicted progress follows elapsed time on successful in-distribution (ID) and OOD trajectories, measured by value-order correlation (VOC). \name-Rewarder ranks trajectories by the expected outcome reward and derives progress by accumulating its dense reward. In OOD ranking, it trails only GPT-6 Astra at 72.0 versus 82.3. It ties Astra at 98.9 ID VOC and reaches 99.6 OOD VOC, exceeding the reported specialized reward baselines on their evaluated metrics. VOC measures ordering on successful episodes and does not establish failure-detection accuracy.

\textbf{Dense reward.} Public reward benchmarks score whole episodes or track task progress. They do not test whether a model can find the moments when a robot acts poorly or fails. DenseRewardBench contains 40 held-out bimanual episodes with human action-quality labels, scored on 4,861 time steps. \name-Rewarder achieves the best value on all four metrics in Table~\ref{tab:dense-reward}, with a macro F1 of 46.2. Its largest advantage is failure detection: it recalls 53.3\% of failure steps, whereas no baseline exceeds 11.7\%. GPT-6 Astra comes closest in bad-action F1 but detects no failures. Higher failure recall has a cost: \name-Rewarder also labels about 11\% of normal steps as failures.

\textbf{Rubric-derived reward.} RubricBench is an in-house evaluation benchmark which contains 40 teleoperated robot episodes from 10 tasks, each graded by humans against a human-created rubric. The metric is the L1 distance between the ground-truth scores and the model predicted scores. \name-Rewarder reaches a pooled L1 of 6.5, which is more accurate than nine of the ten general-purpose models (Table~\ref{tab:rubric}). Astra achieves 3.6. These are system-level comparisons. Appendix~\ref{sec:reward-eval-protocol} specifies model-specific camera settings, missing-response accounting, and further details.

\subsection{Annotation Efficiency}
\label{sec:efficiency}

Figure~\ref{fig:annotation-efficiency} compares \name with six proprietary models at three reasoning settings on a fixed 50-episode subset of DenseCaptionBench, comprising 20 human, 15 parallel-gripper, and 15 dexterous-hand episodes. Quality equally weights segmentation, reference-window captioning, and joint annotation across the three embodiments. \name achieves 57.1\% quality at an estimated \$0.72 per hour of source video under an eight-GB300 deployment at \$10/GPU-hour. GPT-6 Astra at maximum reasoning effort achieves 53.7\% quality at a known-usage API cost of \$88.49 per hour. The estimated cost is approximately $125\times$ lower, with mean quality 3.3 percentage points above Astra, computed from unrounded means. Cost accounting, reference versions, confidence intervals, and quota-based throughput estimates appear in Appendix~\ref{sec:annotation-cost-estimation}.

\section{Qualitative Examples}
\label{sec:annotation-qualitative}

\begin{figure}[t]
  \centering
  \includegraphics[width=\linewidth]{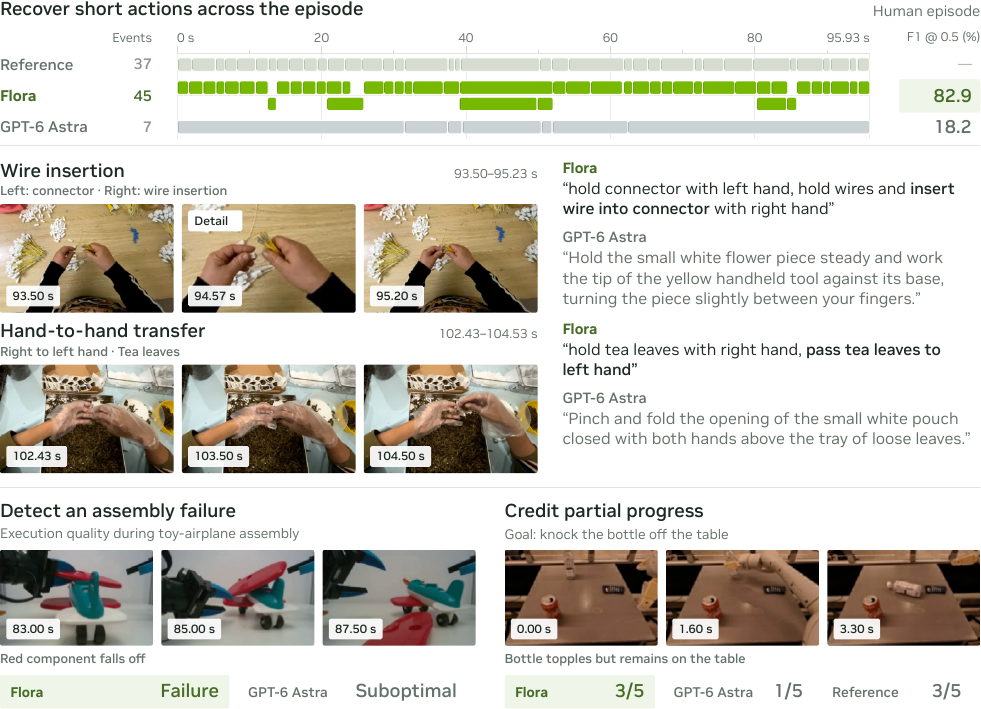}
  \caption{\textbf{From intended actions to execution quality.} Selected examples show temporal coverage (top), human action captions (middle), and failure recognition and partial progress on DenseRewardBench and RoboRewardBench (bottom).}
  \label{fig:qualitative-capabilities}
  \label{fig:annotation-qualitative}
  \label{fig:reward-qualitative}
\end{figure}

Figure~\ref{fig:qualitative-capabilities} compares \name to GPT-6 Astra on temporal coverage, intended actions, and reward decisions. Additional qualitative evidence appears in Appendix~\ref{sec:additional-qualitative}.

\textbf{Temporal coverage and local operations.} In the selected episode, \name matches 34 of 37 reference intervals, compared with four for Astra, yielding F1@0.5 scores of 82.9 and 18.2. The caption examples show the importance of preserving the intended operation: \name identifies wire insertion and a right-to-left handover, where Astra describes a flower piece and pouch folding, respectively. The first caption also distinguishes the hand holding the connector from the hand inserting the wire.

\textbf{Failure severity and partial outcomes.} Both models describe the falling component, but only \name-Rewarder assigns the failure label. In the bottle task, \name-Rewarder distinguishes partial progress from complete failure: toppling the bottle without knocking it off the table receives the reference score of 3/5, whereas Astra assigns 1/5.

\Needspace{6\baselineskip}

\section{Methods}
\label{sec:method}

We now go into technical details of how \name is created. \name aligns intended-action descriptions with execution quality. Captioning localizes each action and describes it using visual context. Reward labeling evaluates individual steps, episode completion, or a user-defined rubric.

\begin{figure}[t]
  \centering
  \includegraphics[width=\linewidth]{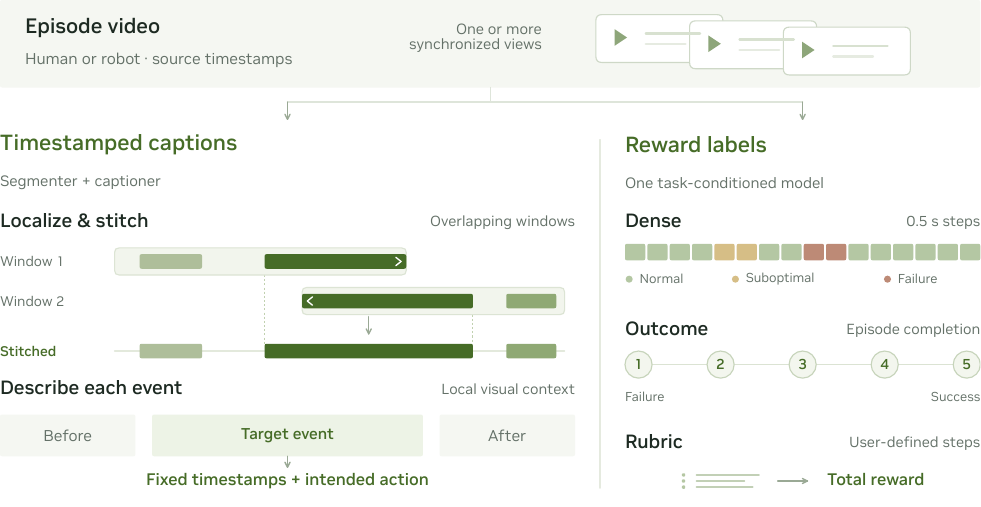}
  \caption{\textbf{From video to reusable supervision.} Overlapping windows are stitched into event intervals for contextual captioning. One reward model provides dense, outcome, or rubric labels. Timelines are schematic.}
  \label{fig:engine-overview}
\end{figure}

\Needspace{12\baselineskip}
\subsection{Timestamped Captions}
\label{sec:textual-annotation}

In robotics we often need dense, timestamped captions. Given a human or robot video $X$ of duration $T$, we output tuples $(s,e,c)$. The string $c$ describes the intended action between time $s$ and time $e$. An action may span several movements and need not succeed. One model provides temporal segments $(s_i,e_i)$ while another provides the captions $c_i$. \name thus produces
\begin{equation}
  \mathcal{A}(X)=\{(s_i,e_i,c_i)\}_{i=1}^{N},
  \qquad I_i=[s_i,e_i),\quad 0\leq s_i<e_i\leq T.
  \label{eq:textual-annotations}
\end{equation}
The pipeline applies a segmentation model to overlapping windows, stitches its predictions into episode-level intervals, and applies a captioning model to each interval with local context (Figure~\ref{fig:engine-overview}).

\Needspace{3\baselineskip}
\textbf{Temporal segmentation and stitching.} We divide an episode into overlapping windows, each with a central output region and surrounding context. The segmentation model predicts start and end times relative to the window, with binary flags indicating continuation past its left and right edges. A deterministic filter retains only fragments intersecting the central region. An action may appear partially in several windows. We convert local timestamps to episode time and join compatible fragments using temporal overlap and predicted continuation flags, while retaining separate intervals for concurrent actions. Boundary refinement reuses the segmenter on shifted windows. Appendix~\ref{sec:annotation-implementation} details the stitching and refinement rules.

\Needspace{6\baselineskip}
\textbf{Contextual captioning.} For each reconstructed interval $I_i$, the pipeline prepares target frames $X_i$ and context frames $X_i^-$ and $X_i^+$ from the nearest non-overlapping events before and after it. The captioning model $g_\phi$ independently generates an instruction $c_i$:
\begin{equation}
  c_i=g_\phi(X_i^-,X_i,X_i^+,q_{\mathrm{cap}}).
  \label{eq:segment-caption}
\end{equation}
The pipeline explicitly labels the before, target, and after blocks. Neighboring frames help resolve object identity and local intent, while the caption remains restricted to $I_i$. Unavailable context is omitted. For multi-camera recordings, the pipeline tiles synchronized views into one image at each sampled time. The prompt identifies camera roles and requests one description of each action across views. Single-camera recordings use the same temporal blocks. The prompt requests a self-contained imperative describing the intended action, objects, manipulator roles, and relevant spatial relations. Failed attempts retain their intended goal without implying success. Finally, the pipeline pairs each generated caption $c_i$ with its fixed interval $I_i$ to form $\mathcal{A}(X)$.

\subsection{Reward Labels}
\label{sec:reward-modeling}

Robot foundation models need rewards at several granularities. \emph{Outcome} rewards judge whether an episode accomplished its task, for deployment monitoring, data filtering, and policy evaluation. \emph{Dense} rewards localize where an episode goes wrong, for per-step credit assignment in RL post-training and fine-grained data curation. For flexible reward modeling at test time, we also support \emph{rubric-derived} rewards. \name-Rewarder is a single model $r_\psi$ that provides all three, selected by its prompt. Outcome and dense prompts tile synchronized camera views into one mosaic per sampled time and condition on the global task instruction. Rubric grading receives separate synchronized camera videos.

\textbf{Outcome reward.}
\label{sec:outcome-reward}
Reinforcement learning for robot policies often learns only from whether each episode succeeded. Simulators report this outcome automatically~\citep{li2025simplevlarl}, but on real robots a person labels each episode~\citep{intelligence2025pistar06}. An outcome reward model can provide these labels instead. The outcome reward scores task completion from complete failure ($s=1$) to complete success ($s=5$), optionally after a short rationale. For a continuous reward, we read the distribution over the five score tokens and rescale its expectation to $[0,1]$ (Eq.~\eqref{eq:outcome-expectation}).

\textbf{Dense execution quality.}
\label{sec:dense-reward}
Feedback at every step helps where a single outcome is too coarse. Each 0.5-second step is labeled failure ($1$), suboptimal ($2$), or normal ($3$). A failure causes a major task failure, an unsafe or destructive event, or a loss of task viability. A suboptimal action is hesitant, inefficient, imprecise, or locally incorrect, but recoverable. Labels are retrospective: later frames may reveal a mistake, but a later recovery does not improve an earlier step. The model also returns a rationale for each contiguous non-normal span. Windows span 16 seconds and advance by 10 seconds. We drop the last four seconds of each non-final window so that retained labels have later visual evidence, then stitch labels on the episode clock. These retrospective annotations are intended for offline labeling.

\Needspace{9\baselineskip}
\textbf{Rubric-derived reward.}
\label{sec:rubric-reward}
A scoring rubric lets users flexibly define rewards at test time. A rubric lists task steps, each with a set of possible scores whose maxima sum to one. Given the episode, instruction, and rubric, the model reasons in thinking mode and then awards one score per step. The total credit is
\begin{equation}
  R_\rho(X)=\sum_{j=1}^{n}\hat y_j,\qquad
  \hat y_j\in V_j,\qquad \sum_{j=1}^{n}\max V_j=1,
  \label{eq:rubric-total}
\end{equation}
where $V_j$ contains the allowed grades for step $j$. Table~\ref{tab:rubric-example} gives an example. Appendix~\ref{sec:reward-implementation} specifies the inputs, output formats, and score extraction.

\subsection{Training}
\label{sec:training}

\textbf{Training infrastructure.} To build the \name data engine, we fine-tune open-source VLMs from the Qwen3.5 family~\citep{qwen3.5}. SFT uses NeMo AutoModel~\citep{nemo-automodel}, and RL uses a custom implementation in PyTorch. Optimization and annotation-RL details appear in Appendices~\ref{sec:annotation-implementation} and~\ref{sec:reward-implementation}.

\textbf{Data strategy.} We train on a combination of open-source datasets, licensed proprietary datasets, and in-house datasets that we collect and label ourselves. Human annotators provide dense, timestamped captions and action-quality labels, guided by detailed labeling guidelines. Reward supervision also includes episode outcomes and rubric grades.

\textbf{Supervised fine-tuning.}
\label{sec:training-textual-annotation}
We initialize the segmentation model $f_\theta$ from Qwen3.5-9B, and the captioning model $g_\phi$ and reward model $r_\psi$ from Qwen3.5-27B. All three models minimize output-token cross-entropy. The segmenter uses LoRA with a frozen visual encoder. The captioner updates the language model and visual merger. Segmentation targets encode window-local intervals and continuation flags. Instruction targets use annotated event boundaries and neighboring context.

\label{sec:training-reward-modeling}
For the reward model, we fine-tune the language model, visual merger, and visual encoder, using a lower learning rate for the visual encoder. Rows from all three reward families are shuffled together, without reweighting or oversampling, and trained jointly for one epoch. Rubric examples are answered in thinking mode, while outcome and dense examples are answered directly. The released reward model uses SFT only.

\textbf{Reinforcement learning for segmentation.} We further train the SFT segmenter with GRPO~\citep{shao2024deepseekmath}. A verifiable reward is
$r_{\mathrm{seg}}=0.45F_{0.5}+0.15F_{0.75}+0.25P_{0.5}+0.15B$,
where $F_\tau$ and $P_\tau$ are temporal F1 and precision at tIoU threshold $\tau$, and $B$ measures endpoint and continuation-flag accuracy. We update only LoRA parameters, with group-mean advantages and KL regularization to the fixed SFT policy. Scoring is restricted to each emission core.

\textbf{Reinforcement learning for captioning.} We further optimize the SFT captioner with GRPO using a reference-grounded point-quality reward. Source-video evidence and reference requirements define
$r_{\mathrm{cap}}=2M/(R+M+F)$,
where $R$ is supported reference mass, $M$ is covered mass, and $F$ penalizes demonstrated factual errors. The reward gives partial credit for supported content. We use group-mean advantages and KL regularization to the fixed SFT policy. Validation selects the two annotation checkpoints independently before benchmark evaluation.

\Needspace{10\baselineskip}
\section{Conclusion}
\label{sec:conclusion}

We present \name, an industrial-quality data engine for timestamped captions, process rewards, and outcome rewards for robotics datasets. \name has been used to train GR00T models. The model is competitive with GPT-6 Astra on captioning benchmarks while improving on reported specialized reward models on RoboReward and Robometer. Its estimated labeling cost offers a practical route to annotating large pretraining corpora. By open-sourcing \name, we hope to enable the community to scale custom robot datasets and to explore novel research directions such as physical auto-research and human-to-robot transfer.

\textbf{Acknowledgements.} We are grateful for discussions and help from Cristaldo Campos, Omar Laymoun, Anita Agasaveeran, Marina Davila, Jenna Diamond, and the broader NVIDIA community.

\par
\bibliographystyle{unsrtnat}
\begingroup
\small\linespread{1.05}\selectfont
\setlength{\bibsep}{4pt}
\raggedright
\bibliography{main}

\begin{thebibliography}{55}
\providecommand{\natexlab}[1]{#1}
\providecommand{\url}[1]{\texttt{#1}}
\expandafter\ifx\csname urlstyle\endcsname\relax
  \providecommand{\doi}[1]{doi: #1}\else
  \providecommand{\doi}{doi: \begingroup \urlstyle{rm}\Url}\fi

\bibitem[Brohan et~al.(2022)Brohan, Brown, Carbajal, Chebotar,
  et~al.]{brohan2022rt1}
Anthony Brohan, Noah Brown, Justice Carbajal, Yevgen Chebotar, et~al.
\newblock {RT-1}: Robotics transformer for real-world control at scale.
\newblock \emph{arXiv preprint arXiv:2212.06817}, 2022.

\bibitem[Zitkovich et~al.(2023)Zitkovich, Yu, Xu, Xu, et~al.]{zitkovich2023rt2}
Brianna Zitkovich, Tianhe Yu, Sichun Xu, Peng Xu, et~al.
\newblock {RT-2}: Vision-language-action models transfer web knowledge to
  robotic control.
\newblock In \emph{Conference on Robot Learning (CoRL)}, 2023.

\bibitem[{Physical Intelligence} et~al.(2025){Physical Intelligence}, Black,
  Brown, et~al.]{intelligence2025pi05}
{Physical Intelligence}, Kevin Black, Noah Brown, et~al.
\newblock $\pi_{0.5}$: a vision-language-action model with open-world
  generalization.
\newblock \emph{arXiv preprint arXiv:2504.16054}, 2025.

\bibitem[Bjorck et~al.(2025)Bjorck, Casta{\~n}eda, Cherniadev, Da,
  et~al.]{bjorck2025gr00t}
Johan Bjorck, Fernando Casta{\~n}eda, Nikita Cherniadev, Xingye Da, et~al.
\newblock {GR00T N1}: An open foundation model for generalist humanoid robots.
\newblock \emph{arXiv preprint arXiv:2503.14734}, 2025.

\bibitem[{Gemini Robotics Team}(2025)]{geminirobotics2025}
{Gemini Robotics Team}.
\newblock Gemini robotics: Bringing {AI} into the physical world.
\newblock \emph{arXiv preprint arXiv:2503.20020}, 2025.

\bibitem[Xiao et~al.(2022)Xiao, Chan, Sermanet, Wahid, et~al.]{xiao2022dial}
Ted Xiao, Harris Chan, Pierre Sermanet, Ayzaan Wahid, et~al.
\newblock Robotic skill acquisition via instruction augmentation with
  vision-language models.
\newblock \emph{arXiv preprint arXiv:2211.11736}, 2022.

\bibitem[Hejna et~al.(2025)Hejna, Bhateja, Jiang, Pertsch, and
  Sadigh]{hejna2024remix}
Joey Hejna, Chethan~Anand Bhateja, Yichen Jiang, Karl Pertsch, and Dorsa
  Sadigh.
\newblock {ReMix}: Optimizing data mixtures for large scale imitation learning.
\newblock In \emph{Conference on Robot Learning (CoRL)}, volume 270 of
  \emph{Proceedings of Machine Learning Research}, pages 145--164, 2025.
\newblock URL \url{https://proceedings.mlr.press/v270/hejna25a.html}.
\newblock CoRL 2024, proceedings published in 2025.

\bibitem[Xie et~al.(2023)Xie, Pham, Dong, Du, et~al.]{xie2023doremi}
Sang~Michael Xie, Hieu Pham, Xuanyi Dong, Nan Du, et~al.
\newblock {DoReMi}: Optimizing data mixtures speeds up language model
  pretraining.
\newblock In \emph{Advances in Neural Information Processing Systems
  (NeurIPS)}, 2023.

\bibitem[Li et~al.(2025)]{li2025simplevlarl}
Haozhan Li et~al.
\newblock {SimpleVLA-RL}: Scaling {VLA} training via reinforcement learning.
\newblock \emph{arXiv preprint arXiv:2509.09674}, 2025.

\bibitem[{Physical Intelligence}(2025)]{intelligence2025pistar06}
{Physical Intelligence}.
\newblock $\pi^*_{0.6}$: a {VLA} that learns from experience.
\newblock \emph{arXiv preprint arXiv:2511.14759}, 2025.

\bibitem[Uesato et~al.(2022)Uesato, Kushman, Kumar, Song,
  et~al.]{uesato2022solving}
Jonathan Uesato, Nate Kushman, Ramana Kumar, Francis Song, et~al.
\newblock Solving math word problems with process- and outcome-based feedback.
\newblock \emph{arXiv preprint arXiv:2211.14275}, 2022.

\bibitem[Lightman et~al.(2024)Lightman, Kosaraju, Burda, Edwards,
  et~al.]{lightman2023verify}
Hunter Lightman, Vineet Kosaraju, Yuri Burda, Harrison Edwards, et~al.
\newblock Let's verify step by step.
\newblock In \emph{International Conference on Learning Representations
  (ICLR)}, 2024.

\bibitem[Duan et~al.(2024)Duan, Pumacay, Kumar, Wang, et~al.]{duan2024aha}
Jiafei Duan, Wilbert Pumacay, Nishanth Kumar, Yi~Ru Wang, et~al.
\newblock {AHA}: A vision-language-model for detecting and reasoning over
  failures in robotic manipulation.
\newblock \emph{arXiv preprint arXiv:2410.00371}, 2024.

\bibitem[Atreya et~al.(2025)Atreya, Pertsch, et~al.]{atreya2025roboarena}
Pranav Atreya, Karl Pertsch, et~al.
\newblock {RoboArena}: Distributed real-world evaluation of generalist robot
  policies.
\newblock \emph{arXiv preprint arXiv:2506.18123}, 2025.

\bibitem[Lu et~al.(2024)Lu, Lu, Lange, Foerster, Clune, and
  Ha]{lu2024aiscientist}
Chris Lu, Cong Lu, Robert~Tjarko Lange, Jakob Foerster, Jeff Clune, and David
  Ha.
\newblock The {AI} scientist: Towards fully automated open-ended scientific
  discovery.
\newblock \emph{arXiv preprint arXiv:2408.06292}, 2024.

\bibitem[Kareer et~al.(2024)Kareer, Patel, Punamiya, Mathur,
  et~al.]{kareer2024egomimic}
Simar Kareer, Dhruv Patel, Ryan Punamiya, Pranay Mathur, et~al.
\newblock {EgoMimic}: Scaling imitation learning via egocentric video.
\newblock \emph{arXiv preprint arXiv:2410.24221}, 2024.

\bibitem[{Macrodata Labs}(2026{\natexlab{a}})]{macrodata2026wgo}
{Macrodata Labs}.
\newblock Segmenting robot video into actionable subtasks.
\newblock \url{https://macrodata.co/blog/annotating-robot-video-subtasks},
  2026{\natexlab{a}}.
\newblock June 25, 2026; accessed September 15, 2026.

\bibitem[{Macrodata Labs}(2026{\natexlab{b}})]{macrodata2026refiner}
{Macrodata Labs}.
\newblock {Refiner}.
\newblock \url{https://github.com/macrodata-labs/refiner}, 2026{\natexlab{b}}.
\newblock Open-source data processing framework. Accessed October 14, 2026.

\bibitem[Palma et~al.(2026)Palma, Kooijmans, Pascal, Meftah, Ellerbach, Russi,
  Bartnik, Rabault, and Wolf]{lerobot2026v060}
Steven Palma, Pepijn Kooijmans, Caroline Pascal, Khalil Meftah, Maxime
  Ellerbach, Martino Russi, Nikodem Bartnik, Nicolas Rabault, and Thomas Wolf.
\newblock {LeRobot} v0.6.0: Imagine, evaluate, improve.
\newblock \url{https://huggingface.co/blog/lerobot-release-v060}, 2026.

\bibitem[{Twelve Labs}(2026)]{twelvelabs2026pegasus15}
{Twelve Labs}.
\newblock {Pegasus} 1.5.
\newblock Model documentation.
  \url{https://docs.twelvelabs.io/v1.3/docs/concepts/models/pegasus/pegasus-1-5},
  2026.
\newblock Accessed October 14, 2026.

\bibitem[Lee et~al.(2026)]{lee2026roboreward}
Tony Lee et~al.
\newblock {RoboReward}: General-purpose vision-language reward models for
  robotics.
\newblock \emph{arXiv preprint arXiv:2601.00675}, 2026.

\bibitem[Liang et~al.(2026)Liang, Korkmaz, Zhang, Hwang, Anwar, Kaushik, Shah,
  Huang, Zettlemoyer, Fox, Xiang, Li, Bobu, Gupta, Tu, Biyik, and
  Zhang]{robometer2026}
Anthony Liang, Yigit Korkmaz, Jiahui Zhang, Minyoung Hwang, Abrar Anwar,
  Sidhant Kaushik, Aditya Shah, Alex~S. Huang, Luke Zettlemoyer, Dieter Fox,
  Yu~Xiang, Anqi Li, Andreea Bobu, Abhishek Gupta, Stephen Tu, Erdem Biyik, and
  Jesse Zhang.
\newblock {Robometer}: Scaling general-purpose robotic reward models via
  trajectory comparisons.
\newblock \emph{arXiv preprint arXiv:2603.02115}, 2026.

\bibitem[{Qwen Team}(2026)]{qwen3.5}
{Qwen Team}.
\newblock {Qwen3.5}: Towards native multimodal agents, February 2026.
\newblock URL \url{https://qwen.ai/blog?id=qwen3.5}.

\bibitem[{NVIDIA}(2025--2026)]{nemo-automodel}
{NVIDIA}.
\newblock {NeMo AutoModel}: {DTensor}-native {SPMD} library for scalable and
  efficient training.
\newblock \url{https://github.com/NVIDIA-NeMo/Automodel}, 2025--2026.
\newblock GitHub repository.

\bibitem[Shao et~al.(2024)Shao, Wang, Zhu, Xu, Song, Bi, Zhang, Zhang, Li, Wu,
  and Guo]{shao2024deepseekmath}
Zhihong Shao, Peiyi Wang, Qihao Zhu, Runxin Xu, Junxiao Song, Xiao Bi, Haowei
  Zhang, Mingchuan Zhang, Y.~K. Li, Y.~Wu, and Daya Guo.
\newblock {DeepSeekMath}: Pushing the limits of mathematical reasoning in open
  language models.
\newblock \emph{arXiv preprint arXiv:2402.03300}, 2024.
\newblock URL \url{https://arxiv.org/abs/2402.03300}.

\bibitem[{OpenAI}(2026{\natexlab{a}})]{openai2026gpt6astra}
{OpenAI}.
\newblock {GPT-6 Astra} system card.
\newblock \url{https://deploymentsafety.openai.com/gpt-6-astra}, September
  2026{\natexlab{a}}.
\newblock OpenAI Deployment Safety Hub.

\bibitem[{OpenAI}(2026{\natexlab{b}})]{openai2026gpt56}
{OpenAI}.
\newblock {GPT-5.6} system card.
\newblock \url{https://deploymentsafety.openai.com/gpt-5-6}, July
  2026{\natexlab{b}}.
\newblock Covers GPT-5.6 Sol, Terra, and Luna. OpenAI Deployment Safety Hub.

\bibitem[{Anthropic}(2026{\natexlab{a}})]{anthropic2026claudeopus5}
{Anthropic}.
\newblock System card: {Claude Opus 5}.
\newblock \url{https://anthropic.com/claude-opus-5-system-card}, July
  2026{\natexlab{a}}.

\bibitem[{Anthropic}(2026{\natexlab{b}})]{anthropic2026claudesonnet5}
{Anthropic}.
\newblock System card: {Claude Sonnet 5}.
\newblock \url{https://anthropic.com/claude-sonnet-5-system-card}, June
  2026{\natexlab{b}}.

\bibitem[{Google DeepMind}(2026{\natexlab{a}})]{google2026gemini31pro}
{Google DeepMind}.
\newblock {Gemini 3.1 Pro} model card.
\newblock \url{https://deepmind.google/models/model-cards/gemini-3-1-pro/},
  February 2026{\natexlab{a}}.

\bibitem[{Google DeepMind}(2026{\natexlab{b}})]{google2026gemini38flash}
{Google DeepMind}.
\newblock {Gemini 3.8 Flash} model card.
\newblock \url{https://deepmind.google/models/model-cards/gemini-3-8-flash/},
  September 2026{\natexlab{b}}.

\bibitem[{Kimi Team}(2026{\natexlab{a}})]{kimiteam2026kimik3}
{Kimi Team}.
\newblock {Kimi K3}: Open frontier intelligence.
\newblock \emph{arXiv preprint arXiv:2607.24653}, 2026{\natexlab{a}}.
\newblock URL \url{https://arxiv.org/abs/2607.24653}.

\bibitem[Tan et~al.(2025)Tan, Chen, Xu, Wang, Ji, Chi, Lyu, Zhao, Chen, Co,
  Xie, Yao, Wang, Wang, and Zhang]{tan2025robodopamine}
Huajie Tan, Sixiang Chen, Yijie Xu, Zixiao Wang, Yuheng Ji, Cheng Chi, Yaoxu
  Lyu, Zhongxia Zhao, Xiansheng Chen, Peterson Co, Shaoxuan Xie, Guocai Yao,
  Pengwei Wang, Zhongyuan Wang, and Shanghang Zhang.
\newblock {Robo-Dopamine}: General process reward modeling for high-precision
  robotic manipulation.
\newblock \emph{arXiv preprint arXiv:2512.23703}, 2025.

\bibitem[Huang et~al.(2026)Huang, Zhang, Hou, Huang, Su, Guo, Lu, Tang, Yang,
  Wang, Peng, Chen, Zhao, and Li]{huang2026rynnvalue}
Dongchi Huang, Hongyin Zhang, Bohan Hou, Siteng Huang, Zhian Su, Hang Guo, Tong
  Lu, Jiahao Tang, Jianfei Yang, Donglin Wang, Peixi Peng, Mingxiu Chen, Deli
  Zhao, and Xin Li.
\newblock {RynnValue}: Scaling robotic value foundation models with temporal
  distance.
\newblock \emph{arXiv preprint arXiv:2608.09853}, 2026.

\bibitem[Kwok et~al.(2026)Kwok, Li, Atreya, Liu, Jiang, Finn, Pavone, Stoica,
  and Mirhoseini]{kwok2026llmverifier}
Jacky Kwok, Shulu Li, Pranav Atreya, Yuejiang Liu, Yixing Jiang, Chelsea Finn,
  Marco Pavone, Ion Stoica, and Azalia Mirhoseini.
\newblock {LLM-as-a-Verifier}: A general-purpose verification framework.
\newblock \emph{arXiv preprint arXiv:2607.05391}, 2026.

\bibitem[{Google DeepMind}(2026{\natexlab{c}})]{google2026gemini35flash}
{Google DeepMind}.
\newblock {Gemini 3.5 Flash}: Model evaluation -- approach, methodology and
  results.
\newblock
  \url{https://storage.googleapis.com/deepmind-media/gemini/gemini_3-5_flash_model_evaluation.pdf},
  2026{\natexlab{c}}.

\bibitem[Levin(2025)]{scale2025physicalai}
Ben Levin.
\newblock Expanding our data engine for physical {AI}.
\newblock \url{https://scale.com/blog/physical-ai}, 2025.

\bibitem[{Macrodata Labs}(2026{\natexlab{c}})]{wgobench2026}
{Macrodata Labs}.
\newblock {WGO-Bench}: What's going on benchmark.
\newblock \url{https://huggingface.co/datasets/macrodata/WGO-Bench},
  2026{\natexlab{c}}.

\bibitem[Cadene et~al.(2024)Cadene, Alibert, Soare, Gallouedec, Zouitine,
  Palma, Kooijmans, Aractingi, Shukor, Aubakirova, Russi, Capuano, Pascal,
  Choghari, Meftah, Ellerbach, Moss, and Wolf]{cadene2024lerobot}
Remi Cadene, Simon Alibert, Alexander Soare, Quentin Gallouedec, Adil Zouitine,
  Steven Palma, Pepijn Kooijmans, Michel Aractingi, Mustafa Shukor, Dana
  Aubakirova, Martino Russi, Francesco Capuano, Caroline Pascal, Jade Choghari,
  Khalil Meftah, Maxime Ellerbach, Jess Moss, and Thomas Wolf.
\newblock {LeRobot}: State-of-the-art machine learning for real-world robotics
  in {PyTorch}.
\newblock \url{https://github.com/huggingface/lerobot}, 2024.

\bibitem[Jung et~al.(2024)Jung, Go, Yi, Jang, et~al.]{twelvelabs2024pegasus}
Raehyuk Jung, Hyojun Go, Jaehyuk Yi, Jiho Jang, et~al.
\newblock {Pegasus-v1} technical report.
\newblock \emph{arXiv preprint arXiv:2404.14687}, 2024.
\newblock URL \url{https://arxiv.org/abs/2404.14687}.

\bibitem[Zawalski et~al.(2024)Zawalski, Chen, Pertsch, Mees, Finn, and
  Levine]{zawalski2024ecot}
Micha{\l} Zawalski, William Chen, Karl Pertsch, Oier Mees, Chelsea Finn, and
  Sergey Levine.
\newblock Robotic control via embodied chain-of-thought reasoning.
\newblock \emph{arXiv preprint arXiv:2407.08693}, 2024.

\bibitem[{Gemma Team}(2025)]{gemma2025gemma3}
{Gemma Team}.
\newblock Gemma 3 technical report.
\newblock \emph{arXiv preprint arXiv:2503.19786}, 2025.

\bibitem[Farabet and Lacombe(2026)]{gemma2026gemma4}
Clement Farabet and Olivier Lacombe.
\newblock Gemma 4: Byte for byte, the most capable open models.
\newblock
  \url{https://blog.google/innovation-and-ai/technology/developers-tools/gemma-4/},
  2026.

\bibitem[Bai et~al.(2025)Bai, Cai, Chen, Chen, et~al.]{bai2025qwen3vl}
Shuai Bai, Yuxuan Cai, Ruizhe Chen, Keqin Chen, et~al.
\newblock {Qwen3-VL} technical report.
\newblock \emph{arXiv preprint arXiv:2511.21631}, 2025.
\newblock URL \url{https://arxiv.org/abs/2511.21631}.

\bibitem[Wang et~al.(2025)]{wang2025internvl35}
Weiyun Wang et~al.
\newblock {InternVL3.5}: Advancing open-source multimodal models in
  versatility, reasoning, and efficiency.
\newblock \emph{arXiv preprint arXiv:2508.18265}, 2025.

\bibitem[{GLM-V Team}(2025)]{glm2025glm45v}
{GLM-V Team}.
\newblock {GLM-4.5V} and {GLM-4.1V-Thinking}: Towards versatile multimodal
  reasoning with scalable reinforcement learning.
\newblock \emph{arXiv preprint arXiv:2507.01006}, 2025.

\bibitem[{Kimi Team}(2025)]{kimi2025kimivl}
{Kimi Team}.
\newblock {Kimi-VL} technical report.
\newblock \emph{arXiv preprint arXiv:2504.07491}, 2025.

\bibitem[{Kimi Team}(2026{\natexlab{b}})]{kimi2026k25}
{Kimi Team}.
\newblock {Kimi K2.5}: Visual agentic intelligence.
\newblock \emph{arXiv preprint arXiv:2602.02276}, 2026{\natexlab{b}}.

\bibitem[Xing et~al.(2025)Xing, Dong, Zang, Cao, Liang, Huang, Wang, Wu, and
  Lin]{xing2025caprl}
Long Xing, Xiaoyi Dong, Yuhang Zang, Yuhang Cao, Jianze Liang, Qidong Huang,
  Jiaqi Wang, Feng Wu, and Dahua Lin.
\newblock {CapRL}: Stimulating dense image caption capabilities via
  reinforcement learning.
\newblock \emph{arXiv preprint arXiv:2509.22647}, 2025.

\bibitem[Yang et~al.(2026)Yang, Xing, Dong, Zang, Cao, Wang, Zhou, Bu, Liang,
  Huang, Wang, Wu, and Lin]{caprlpp2026}
Penghui Yang, Long Xing, Xiaoyi Dong, Yuhang Zang, Yuhang Cao, Yibin Wang,
  Yujie Zhou, Jiazi Bu, Jianze Liang, Qidong Huang, Jiaqi Wang, Feng Wu, and
  Dahua Lin.
\newblock {CapRL++}: Unified reinforcement learning with verifiable rewards for
  dense image and video captioning.
\newblock \emph{arXiv preprint arXiv:2606.09393}, 2026.

\bibitem[Ma et~al.(2023{\natexlab{a}})Ma, Sodhani, Jayaraman, Bastani, Kumar,
  and Zhang]{ma2022vip}
Yecheng~Jason Ma, Shagun Sodhani, Dinesh Jayaraman, Osbert Bastani, Vikash
  Kumar, and Amy Zhang.
\newblock {VIP}: Towards universal visual reward and representation via
  value-implicit pre-training.
\newblock In \emph{International Conference on Learning Representations
  (ICLR)}, 2023{\natexlab{a}}.

\bibitem[Ma et~al.(2023{\natexlab{b}})Ma, Kumar, Zhang, Bastani, and
  Jayaraman]{ma2023liv}
Yecheng~Jason Ma, Vikash Kumar, Amy Zhang, Osbert Bastani, and Dinesh
  Jayaraman.
\newblock {LIV}: Language-image representations and rewards for robotic
  control.
\newblock In \emph{International Conference on Machine Learning (ICML)},
  2023{\natexlab{b}}.
\newblock URL \url{https://proceedings.mlr.press/v202/ma23b.html}.

\bibitem[Rocamonde et~al.(2024)Rocamonde, Montesinos, Nava, Perez, and
  Lindner]{rocamonde2023vlmrm}
Juan Rocamonde, Victoriano Montesinos, Elvis Nava, Ethan Perez, and David
  Lindner.
\newblock Vision-language models are zero-shot reward models for reinforcement
  learning.
\newblock In \emph{International Conference on Learning Representations
  (ICLR)}, 2024.

\bibitem[Sontakke et~al.(2023)Sontakke, Zhang, Arnold, Pertsch, B{\i}y{\i}k,
  Sadigh, Finn, and Itti]{sontakke2023roboclip}
Sumedh~A. Sontakke, Jesse Zhang, S{\'e}bastien M.~R. Arnold, Karl Pertsch,
  Erdem B{\i}y{\i}k, Dorsa Sadigh, Chelsea Finn, and Laurent Itti.
\newblock {RoboCLIP}: One demonstration is enough to learn robot policies.
\newblock In \emph{Advances in Neural Information Processing Systems
  (NeurIPS)}, 2023.

\bibitem[Ma et~al.(2024)Ma, Hejna, Wahid, Fu, et~al.]{ma2024gvl}
Yecheng~Jason Ma, Joey Hejna, Ayzaan Wahid, Chuyuan Fu, et~al.
\newblock Vision language models are in-context value learners.
\newblock \emph{arXiv preprint arXiv:2411.04549}, 2024.

\end{thebibliography}
\endgroup

\clearpage
\appendix
\section{Complete Benchmark Results}
\label{sec:complete-results}
Figure~\ref{fig:release-capabilities} and Table~\ref{tab:release-scorecard} summarize selected baselines. The tables below retain every evaluated model and metric breakdown. All values are scaled by 100. Green identifies \name, and shaded cells mark the best displayed result, including ties. Evaluation protocols appear in Appendices~\ref{sec:annotation-protocol} and~\ref{sec:reward-eval-protocol}.

\subsection{Timestamped captions: WGO-Bench}

\begin{table}[H]
  \centering
  \floraTableStyle
  \setlength{\tabcolsep}{1.5pt}
  \caption{\textbf{WGO-Bench results.} Subset scores and equal-weight averages (\%, $\uparrow$). Off-the-shelf models use the \hyperref[par:common-pipeline]{common pipeline}, and \name uses its default pipeline (Section~\ref{sec:textual-annotation}).}
  \label{tab:wgo}
  \begin{tabular*}{\linewidth}{@{\extracolsep{\fill}}
    >{\raggedright\arraybackslash}p{0.19\linewidth}
    *{3}{M{0.067\linewidth}M{0.062\linewidth}M{0.07\linewidth}M{0.045\linewidth}}@{}}
    \toprule
    & \multicolumn{4}{c}{\shortstack{\textbf{Segmentation}\\F1@0.75}} &
      \multicolumn{4}{c}{\shortstack{\textbf{Captioning}\\Gold-window accuracy}} &
      \multicolumn{4}{c}{\shortstack{\textbf{Joint annotation}\\Semantic F1@0.75}} \\
    \cmidrule(lr){2-5}\cmidrule(lr){6-9}\cmidrule(l){10-13}
    \textbf{Model / system} & {\fontsize{8}{9}\selectfont HomER} & DROID & {\fontsize{8}{9}\selectfont Galaxea} & \textbf{Avg} &
      {\fontsize{8}{9}\selectfont HomER} & DROID & {\fontsize{8}{9}\selectfont Galaxea} & \textbf{Avg} & {\fontsize{8}{9}\selectfont HomER} & DROID & {\fontsize{8}{9}\selectfont Galaxea} & \textbf{Avg} \\
    \midrule
    \addlinespace[3pt]
    \multicolumn{13}{@{}l}{\textcolor{black!55}{\footnotesize\scshape Proprietary models}} \\
    \addlinespace[1pt]
    GPT-6 Astra & \floraRankFirst{25.2} & \floraRankFirst{65.5} & 47.2 & 46.0 & \floraRankFirst{64.0} & \floraRankFirst{82.0} & \floraRankFirst{86.2} & \floraRankFirst{77.4} & 19.3 & \floraRankFirst{54.5} & 43.1 & \floraRankFirst{39.0} \\
    GPT-5.6 Sol & 23.5 & 44.0 & 33.5 & 33.7 & 48.7 & 62.7 & 58.5 & 56.6 & 16.9 & 33.6 & 23.4 & 24.6 \\
    GPT-5.6 Luna & 20.6 & 32.3 & 37.8 & 30.2 & 46.8 & 55.3 & 57.7 & 53.3 & 13.3 & 22.3 & 24.0 & 19.9 \\
    Gemini 3.8 Flash & 23.9 & 30.8 & 24.5 & 26.4 & 63.0 & 70.0 & 79.7 & 70.9 & \floraRankFirst{20.6} & 23.7 & 20.9 & 21.7 \\
    Gemini 3.1 Pro & 12.0 & 27.0 & 38.1 & 25.7 & 57.7 & 58.7 & 56.9 & 57.8 & 7.5 & 14.9 & 29.4 & 17.3 \\
    Claude Opus 5 & 10.6 & 38.8 & 27.2 & 25.5 & 45.1 & 50.0 & 52.8 & 49.3 & 7.4 & 20.1 & 18.7 & 15.4 \\
    Claude Sonnet 5 & 3.2 & 29.5 & 29.6 & 20.8 & 36.8 & 31.3 & 37.4 & 35.2 & 2.6 & 10.9 & 19.4 & 11.0 \\
    \addlinespace[3pt]
    \multicolumn{13}{@{}l}{\textcolor{black!55}{\footnotesize\scshape Open-weight models}} \\
    \addlinespace[1pt]
    Qwen3.5-27B & 12.7 & 28.8 & 25.7 & 22.4 & 42.3 & 39.3 & 49.6 & 43.7 & 5.4 & 15.8 & 20.0 & 13.7 \\
    Qwen3.5-397B-A17B & 10.5 & 28.7 & 19.4 & 19.5 & 47.0 & 44.7 & 46.3 & 46.0 & 8.2 & 14.0 & 11.5 & 11.2 \\
    Kimi K3 & 18.9 & 28.2 & 31.1 & 26.1 & 44.0 & 57.3 & 50.4 & 50.6 & 12.7 & 17.9 & 19.6 & 16.7 \\
    \midrule

    \textbf{\name (ours)}      & \floraOurScore{23.9} & \floraOurScore{56.1} & \floraOurBest{70.8} & \floraOurBest{50.3} & \floraOurBest{64.0} & \floraOurScore{55.3} & \floraOurScore{74.8} & \floraOurScore{64.7} & \floraOurScore{17.9} & \floraOurScore{32.5} & \floraOurBest{53.1} & \floraOurScore{34.5} \\
    \bottomrule
  \end{tabular*}
\end{table}

\subsection{Timestamped captions: DenseCaptionBench}

\begin{table}[H]
  \centering
  \floraTableStyle
  \setlength{\tabcolsep}{1.5pt}
  \caption{\textbf{DenseCaptionBench results.} Scores on 100 human, 100 parallel-gripper (Grip.), and 100 dexterous-hand (Dex.) episodes under the shared point-quality protocol (\%, $\uparrow$). Off-the-shelf models use the \hyperref[par:common-pipeline]{common pipeline}, and \name uses its production pipeline (Section~\ref{sec:textual-annotation}).}
  \label{tab:internal-e2e}
  \begin{tabular*}{\linewidth}{@{\extracolsep{\fill}}
    >{\raggedright\arraybackslash}p{0.19\linewidth}
    *{3}{M{0.067\linewidth}M{0.062\linewidth}M{0.07\linewidth}M{0.045\linewidth}}@{}}
    \toprule
    \multirow{3}{*}{\textbf{Model / system}} &
      \multicolumn{4}{c}{\shortstack{\textbf{Segmentation}\\F1@0.5}} &
      \multicolumn{4}{c}{\shortstack{\textbf{Oracle captioning}\\Point quality}} &
      \multicolumn{4}{c}{\shortstack{\textbf{Joint annotation}\\Point F1@0.5}} \\
    \cmidrule(lr){2-5}\cmidrule(lr){6-9}\cmidrule(l){10-13}
    & \multirow{2}{*}{Human} & \multicolumn{3}{c}{Robot} &
      \multirow{2}{*}{Human} & \multicolumn{3}{c}{Robot} &
      \multirow{2}{*}{Human} & \multicolumn{3}{c}{Robot} \\
    \cmidrule(lr){3-5}\cmidrule(lr){7-9}\cmidrule(l){11-13}
    & & Grip. & Dex. & \textbf{Avg} & & Grip. & Dex. & \textbf{Avg} &
      & Grip. & Dex. & \textbf{Avg} \\
    \midrule
    \addlinespace[3pt]
    \multicolumn{13}{@{}l}{\textcolor{black!55}{\footnotesize\scshape Proprietary models}} \\
    \addlinespace[1pt]
    GPT-6 Astra & 44.2 & 70.7 & \floraRankFirst{73.1} & \floraRankFirst{71.9} & 60.3 & \floraRankFirst{79.6} & \floraRankFirst{80.8} & \floraRankFirst{80.2} & 29.7 & \floraRankFirst{57.7} & \floraRankFirst{58.1} & \floraRankFirst{57.9} \\
    GPT-5.6 Sol & 32.6 & 32.4 & 59.2 & 45.8 & 36.4 & 68.8 & 70.3 & 69.6 & 14.6 & 22.8 & 44.3 & 33.6 \\
    GPT-5.6 Luna & 21.4 & 23.8 & 51.9 & 37.9 & 28.5 & 54.2 & 52.1 & 53.2 & 8.1 & 13.7 & 31.5 & 22.6 \\
    Gemini 3.8 Flash & 33.5 & 33.7 & 57.9 & 45.8 & 46.7 & 64.4 & 67.5 & 66.0 & 17.7 & 22.6 & 41.7 & 32.2 \\
    Gemini 3.1 Pro & 20.7 & 29.0 & 43.3 & 36.2 & 39.1 & 44.7 & 48.0 & 46.4 & 9.3 & 11.5 & 21.1 & 16.3 \\
    Claude Opus 5 & 22.7 & 42.5 & 49.2 & 45.9 & 31.0 & 50.0 & 49.3 & 49.7 & 8.2 & 23.2 & 24.0 & 23.6 \\
    Claude Sonnet 5 & 13.8 & 33.6 & 42.1 & 37.9 & 23.0 & 31.9 & 40.4 & 36.2 & 4.4 & 13.1 & 17.8 & 15.5 \\
    \addlinespace[3pt]
    \multicolumn{13}{@{}l}{\textcolor{black!55}{\footnotesize\scshape Open-weight models}} \\
    \addlinespace[1pt]
    Qwen3.5-27B & 13.3 & 21.5 & 54.8 & 38.2 & 29.3 & 44.2 & 47.4 & 45.8 & 5.3 & 11.0 & 27.7 & 19.4 \\
    Qwen3.5-397B-A17B & 19.6 & 26.3 & 47.6 & 37.0 & 29.1 & 46.2 & 52.7 & 49.5 & 7.8 & 13.9 & 25.7 & 19.8 \\
    Kimi K3 & 21.1 & 28.7 & 53.6 & 41.2 & 30.3 & 45.6 & 50.3 & 48.0 & 7.9 & 15.0 & 29.4 & 22.2 \\
    \midrule

    \textbf{\name (ours)}      & \floraOurBest{54.6} & \floraOurBest{71.0} & \floraOurScore{62.7} & \floraOurScore{66.9} & \floraOurBest{66.5} & \floraOurScore{74.3} & \floraOurScore{73.1} & \floraOurScore{73.7} & \floraOurBest{37.4} & \floraOurScore{52.6} & \floraOurScore{49.4} & \floraOurScore{51.0} \\
    \bottomrule
  \end{tabular*}
\end{table}

\subsection{Public reward benchmarks}

\begin{table}[H]
  \centering
  \rewardTableStyle
  \caption{\textbf{Public reward benchmarks.} \name-Rewarder has the lowest reported MAE and exceeds the specialized reward baselines on their reported metrics. Published specialized-model scores retain their original protocols. All metrics are multiplied by 100.}
  \label{tab:public-reward}
  \label{tab:roboreward}
  \label{tab:robometer}
  \begin{tabular*}{\linewidth}{@{\extracolsep{\fill}}lrrrr@{}}
    \toprule
    & \multicolumn{1}{c}{\textbf{RoboRewardBench}} &
    \multicolumn{3}{c}{\textbf{RobometerBench}} \\
    \cmidrule(lr){2-2}\cmidrule(l){3-5}
    \textbf{Model} & MAE $\downarrow$ & OOD Kendall $\tau_a$ $\uparrow$ &
    ID VOC $\uparrow$ & OOD VOC $\uparrow$ \\
    \midrule
    \multicolumn{5}{@{}l}{\textcolor{black!55}{\footnotesize\scshape Proprietary models}} \\
    GPT-6 Astra & 55.4 & \floraRankFirst{82.3} & \floraRankFirst{98.9} & \floraRankFirst{99.7} \\
    GPT-5.6 Sol & 78.0 & 64.2 & 97.1 & 99.3 \\
    GPT-5.6 Luna & 77.9 & 59.3 & 95.7 & 98.7 \\
    Gemini 3.8 Flash & 83.3 & 63.3 & 96.9 & 98.6 \\
    Gemini 3.1 Pro & 99.6 & 57.2 & 85.3 & 96.8 \\
    Claude Opus 5 & 88.2 & 63.1 & 91.8 & 98.6 \\
    Claude Sonnet 5 & 96.6 & 48.3 & 86.1 & 89.7 \\
    \addlinespace[3pt]
    \multicolumn{5}{@{}l}{\textcolor{black!55}{\footnotesize\scshape Open-weight general models}} \\
    Qwen3.5-27B & 84.8 & 53.4 & 91.4 & 94.1 \\
    Qwen3.5-397B-A17B & 87.1 & 48.8 & 91.5 & 97.1 \\
    Kimi K3 & 68.2 & 63.5 & 97.2 & \floraRankFirst{99.7} \\
    \addlinespace[3pt]
    \multicolumn{5}{@{}l}{\textcolor{black!55}{\footnotesize\scshape Specialized reward models}} \\
    RoboDopamine & \rewardPending & 45.0 & 79.0 & 80.0 \\
    RynnValue & \rewardPending & 67.5 & \rewardPending & \rewardPending \\
    LLM-as-a-Verifier & 72.0 & \rewardPending & \rewardPending & \rewardPending \\
    RoboReward & 66.5 & 47.0 & 82.0 & 88.0 \\
    Robometer & 75.0 & 65.5 & 92.0 & 94.0 \\
    \midrule
    \multicolumn{5}{@{}l}{\textcolor{black!55}{\footnotesize\scshape Ours}} \\
    \textbf{\name-Rewarder} & \floraOurBest{40.0} & \floraOurScore{72.0} & \floraOurBest{98.9} & \floraOurScore{99.6} \\
    \bottomrule
  \end{tabular*}
\end{table}

\subsection{Dense and rubric rewards}

\begin{table}[H]
  \centering
  \rewardTableStyle
  \caption{\textbf{Internal reward benchmarks.} \name-Rewarder achieves the best value on all four DenseRewardBench metrics and grades rubrics more accurately than nine of the ten general-purpose models. All metrics are multiplied by 100.}
  \label{tab:internal-reward}
  \label{tab:dense-reward}
  \label{tab:rubric}
  \begin{tabular*}{\linewidth}{@{\extracolsep{\fill}}lrrrrr@{}}
    \toprule
    & \multicolumn{4}{c}{\textbf{DenseRewardBench}} &
    \multicolumn{1}{c}{\textbf{RubricBench}} \\
    \cmidrule(lr){2-5}\cmidrule(l){6-6}
    \textbf{Model} & \shortstack{Macro\\F1 $\uparrow$} &
    \shortstack{Balanced\\acc. $\uparrow$} & \shortstack{Bad-action\\F1 $\uparrow$} &
    \shortstack{Failure\\recall $\uparrow$} & \shortstack{Pooled\\L1 $\downarrow$} \\
    \midrule
    \multicolumn{6}{@{}l}{\textcolor{black!55}{\footnotesize\scshape Proprietary models}} \\
    GPT-6 Astra & 40.7 & 44.1 & 44.8 & 0.0 & \floraRankFirst{3.6} \\
    GPT-5.6 Sol & 36.3 & 37.8 & 26.4 & 1.0 & 8.4 \\
    GPT-5.6 Luna & 34.7 & 34.9 & 17.6 & 2.3 & 22.6 \\
    Gemini 3.8 Flash & 38.8 & 40.2 & 31.2 & 11.7 & 32.6 \\
    Gemini 3.1 Pro & 32.8 & 33.2 & 16.0 & 3.0 & 15.1 \\
    Claude Opus 5 & 33.9 & 35.1 & 22.9 & 0.3 & 17.1 \\
    Claude Sonnet 5 & 31.5 & 32.4 & 13.3 & 0.3 & 22.4 \\
    \addlinespace[3pt]
    \multicolumn{6}{@{}l}{\textcolor{black!55}{\footnotesize\scshape Open-weight general models}} \\
    Qwen3.5-27B & 30.8 & 32.4 & 7.7 & 5.7 & 15.4 \\
    Qwen3.5-397B-A17B & 34.7 & 39.2 & 31.3 & 3.3 & 8.9 \\
    Kimi K3 & 31.2 & 32.3 & 17.6 & 0.3 & 15.6 \\
    \midrule
    \multicolumn{6}{@{}l}{\textcolor{black!55}{\footnotesize\scshape Ours}} \\
    \textbf{\name-Rewarder} & \floraOurBest{46.2} & \floraOurBest{54.2} & \floraOurBest{47.7} & \floraOurBest{53.3} & \floraOurScore{6.5} \\
    \bottomrule
  \end{tabular*}
\end{table}

\textbf{Baselines.} We consider both VLM models and full labeling pipelines as baselines. For models we consider publicly available proprietary models: GPT-6 Astra \citep{openai2026gpt6astra}, GPT-5.6 Sol and Luna \citep{openai2026gpt56}, Claude Opus 5 and Sonnet 5 \citep{anthropic2026claudeopus5, anthropic2026claudesonnet5}, Gemini 3.1 Pro \citep{google2026gemini31pro} and Gemini 3.8 Flash \citep{google2026gemini38flash}. For open-weights models we consider Kimi K3 \citep{kimiteam2026kimik3} and Qwen3.5-27B and Qwen3.5-397B-A17B \citep{qwen3.5}. We also compare with a few specialized reward models: RoboDopamine~\citep{tan2025robodopamine}, RynnValue~\citep{huang2026rynnvalue}, LLM-as-a-Verifier~\citep{kwok2026llmverifier}, RoboReward~\citep{lee2026roboreward}, and Robometer~\citep{robometer2026}. Pipeline backbones and implementations appear in Appendix~\ref{sec:wgo-pipelines}.

\FloatBarrier
\section{Timestamped Captioning: Evaluation Details}
\label{sec:annotation-protocol}

We evaluate localization, caption quality on reference intervals, and joint
annotation on complete episodes. The following sections specify benchmark
construction, visual inputs, scoring, and supplementary pipeline comparisons.
All results use the model configurations reported in the main text.

\subsection{DenseCaptionBench construction}
\label{sec:annotation-benchmark-construction}
Table~\ref{tab:annotation-tracks} summarizes the fixed 300-episode DenseCaptionBench.
Every episode supports all three evaluation tasks. The parallel-gripper
track is labeled YAM in Table~\ref{tab:internal-e2e}, and Dex denotes the
dexterous-hand track.

\begin{table}[!htbp]
\centering
\floraTableStyle
\caption{\textbf{DenseCaptionBench composition.} Reference intervals may
include idle gaps and concurrent intended actions.}
\label{tab:annotation-tracks}
\begin{tabularx}{\linewidth}{@{}lXrr@{}}
\toprule
\textbf{Embodiment} & \textbf{Visual evidence} & \textbf{Episodes} & \textbf{References} \\
\midrule
Human & Egocentric camera & 100 & 3,220 \\
Parallel grippers & Main and two wrist cameras & 100 & 1,266 \\
Dexterous hands & Main and two wrist cameras & 100 & 391 \\
\midrule
\textbf{Total} & & \textbf{300} & \textbf{4,877} \\
\bottomrule
\end{tabularx}
\end{table}

\textbf{Coverage-based selection.} We select episodes using available task,
action, environment, and manipulator-coordination metadata. For a feature
group $g$, let $\mathcal{T}_g$ be its labels, $N_t$ the number of eligible
pool episodes with label $t$, and $n_t(S)$ the number in a selected set $S$.
The diminishing-return coverage objective is
\begin{equation}
  \Phi(S)=\sum_{g\in\mathcal{G}}\frac{1}{|\mathcal{T}_g|}
    \sum_{t\in\mathcal{T}_g}\sqrt{\frac{n_t(S)}{N_t}}.
  \label{eq:benchmark-coverage}
\end{equation}
Greedy selection adds the admissible episode with the largest marginal gain,
using fixed-seed tie-breaking and duplicate filtering. Counts measure episode
presence, not repeated event frequency. Robot selection also limits repeated
tasks and collection sessions and maintains source-family quotas. Missing
metadata contributes no coverage gain. Initial selection uses existing
annotations and metadata without model outputs or scores, and imposes no
single-/multiple-event, duration, or event-density quota.

The human track spans 56 environment labels and 95 task labels. The two
robot tracks span 99 and 75 source-qualified task groups, respectively.
These are metadata coverage measures, not independently verified counts of
distinct skills. Unavailable or excluded sources are replaced using the
same metadata-based criteria before comparison on the final roster.
The human cohort is therefore conditioned on source and service availability.

\textbf{Reference construction.} Intended actions originate from human
annotations and are refined through source review and adjudication. Reference
specifications record action requirements, object identities, coordination,
phase dependencies, and observation limits before candidate-caption access
in the scoring stage. Robot references include 1,364 units with established
intended actions and 293 observed-operation proxies. These types remain
separately identified. Sampled review uses temporal coverage and targeted
boundary inspection, with uncertainty retained when the evidence is
insufficient. It does not establish continuous visual coverage or exact
transition times. Reference eligibility is shared across compared models.

\subsection{Inputs and inference settings}
\label{sec:annotation-inputs}

\textbf{Common pipeline.} Off-the-shelf models use the two-stage interface
in Section~\ref{sec:textual-annotation-eval}. One call segments timestamped
episode contact sheets. Each interval is then captioned independently from
labeled before, target, and after blocks, using the same model in both stages.
Oracle captioning substitutes reference intervals for predicted ones.
Generation never receives reference captions, neighboring labels, or
segmentation-stage draft captions. Captioning cannot alter interval boundaries.

For WGO and human episodes, segmentation samples at 2 Hz, capped at 1,024
frames. Caption targets sample at 4 Hz, capped at 256 frames. Context extends
at most 2 seconds per side at 2 Hz, with at most five frames per side.
Capped sampling preserves the full interval span. Frames use timestamped
$320\times240$ letterboxed tiles in $4\times4$ contact sheets, encoded at
JPEG quality 90.

Robot episodes retain their three camera streams. Segmentation samples
uniform temporal bin centers at nominal 2 Hz, capped at 256 mosaics.
Caption targets use duration-adaptive sampling at nominal 8 Hz. Nominal
budgets are 8--40 target frames and 4--12 frames per available context block,
up to 64 frames total. Source availability can reduce these counts.
Context remains limited to 2 seconds per side. Main and wrist views
form $768\times648$ mosaics for the parallel-gripper track and
$640\times720$ mosaics for the dexterous-hand track. A fixed image-scale
ladder fits lossless PNG inputs within the request budget without removing
selected timestamps. Temporal packing may differ across request containers.
Source-limited inputs retain explicit padding or unavailable-input status.

\textbf{Native \name pipeline.} The segmenter samples 64 frames at 4 Hz in
16-second windows, with a 12-second core and 2-second guards. Long-horizon
robot sources use 32-second windows, a 24-second core, and 4-second guards.
Local output coordinates use 20 Hz ticks with continuation flags. The
production parser, adaptive window processing, and stitching are fixed
before evaluation. Inference uses BF16 computation while retaining Gated
DeltaNet parameters and temporal LoRA tensors in FP32. Greedy decoding
allows 1,024 temporal tokens and 512 caption tokens. The captioner uses the duration-adaptive 8 Hz sampling rule
above through its trained video processor. WGO uses adjacent-event context,
while DenseCaptionBench evaluation clips real context to 2 seconds per side.
Generation on DenseCaptionBench receives neither task labels nor reference text.
All WGO systems receive the dataset episode instruction in both stages.
These are complete-system comparisons. Native processors, prompts, window
schedules, and inference compute differ from the common pipeline.

\begin{table}[!htbp]
\centering
\floraTableStyle
\caption{\textbf{Baseline decoding configurations.} Output limits are in
thousands of tokens (k = 1,024). Default denotes an omitted request field,
not a verified reasoning level.}
\label{tab:inhouse-inference-settings}
\begin{tabularx}{\linewidth}{@{}lXcc@{}}
\toprule
& & \multicolumn{2}{c}{\textbf{Output limit}} \\
\cmidrule(l){3-4}
\textbf{Model} & \textbf{Reasoning and sampling} & \textbf{DenseCaptionBench} & \textbf{WGO} \\
\midrule
GPT-6 Astra & \texttt{xhigh}, default sampling & 32k & 32k \\
GPT-5.6 Sol / Luna & \texttt{xhigh}, default sampling & 32k & 32k \\
Gemini 3.8 Flash & Default reasoning, temperature 0 & 32k & 32k \\
Gemini 3.1 Pro & Default reasoning, temperature 0 & 16k & 32k \\
Claude Opus 5 / Sonnet 5 & Provider defaults & 32k & 32k \\
Kimi K3 & \texttt{max}, default sampling & 32k & 32k \\
Qwen3.5-27B & Greedy on DenseCaptionBench, sampling on WGO & 16k & 32k \\
Qwen3.5-397B-A17B & Temperature 0 on DenseCaptionBench, sampling on WGO & 16k & 32k \\
\bottomrule
\end{tabularx}
\end{table}

Qwen WGO runs enable thinking with temperature 1, top-$p=0.95$,
top-$k=20$, and presence penalty 1.5. On DenseCaptionBench, the 27B runs disable thinking
and use greedy decoding. The 27B baseline uses official BF16 weights.
For hosted models, weight revisions, numerical precision, and effective
backend settings are not independently exposed. Requested settings are
fixed within each benchmark and are not compute matched.

\subsection{DenseCaptionBench scoring and aggregation}
\label{sec:annotation-metrics}

\textbf{Temporal segmentation.} Let $\widehat{\mathcal I}$ and $\mathcal I^\star$ denote predicted and
reference intervals. An edge is eligible at tIoU $\geq0.5$. Caption-blind
maximum-cardinality one-to-one matching gives $\mathrm{TP}$ and
\begin{equation}
  F_{\mathrm{seg}}=\frac{2\mathrm{TP}}
    {|\widehat{\mathcal I}|+|\mathcal I^\star|}.
  \label{eq:inhouse-temporal}
\end{equation}
We also retain tIoU $0.75$ as a sensitivity setting. Boundary arithmetic uses
integer microseconds. There is no reference-edge snapping. Only the final
outer endpoint is intersected with the available media clock.

\Needspace{5\baselineskip}
\textbf{Caption quality.} Each reference contains requirements in four groups:
action, identity, coordination, and detail, weighted $0.50$, $0.25$, $0.20$,
and $0.05$. Each supported requirement $a$ has a fixed mass $m_a$ and a
binary coverage indicator $z_a$. Coverage requires a supporting caption
span and satisfaction of the reference's logical dependencies. A
phase-specific requirement additionally requires its associated action-phase
predicates, so a state from another phase cannot receive credit.
With supported requirements $\mathcal A$ and distinct demonstrated
factual-error families $\mathcal E$, we compute
\begin{equation}
  R=\sum_{a\in\mathcal A}m_a,\qquad
  M=\sum_{a\in\mathcal A}m_a z_a,\qquad
  F=1.25\sum_{h\in\mathcal E}w_{g(h)},
  \label{eq:inhouse-point-masses}
\end{equation}
where $w_{g(h)}$ is the weight of error family $h$'s semantic group.
Repeated descriptions of the same erroneous fact incur one penalty.
The resulting quality is
\begin{equation}
  q=\frac{2M}{R+M+F}.
  \label{eq:inhouse-point-quality}
\end{equation}
Equivalent paraphrases receive credit. Unverifiable assertions are distinct
from demonstrated errors, and confidence does not multiply the score.
Source-uncertain requirements are excluded from supported mass. A pair with
no supported mass is unavailable. Failed attempts may still match the
intended action without claiming successful execution.

\textbf{Oracle and joint tasks.} Oracle captioning averages $q$ over all
reference intervals in an episode. Joint annotation first constructs every
temporally eligible prediction--reference edge, then finds the one-to-one
assignment $\mathcal M$ maximizing total quality:
\begin{equation}
  F_{\mathrm{joint}}=
    \frac{2\displaystyle\max_{\mathcal M}
      \sum_{(i,j)\in\mathcal M}q_{ij}}
    {|\widehat{\mathcal I}|+|\mathcal I^\star|}.
  \label{eq:inhouse-joint}
\end{equation}
Empty captions retain their predicted intervals in the denominator.
Joint F1 is not the product of separately averaged segmentation and caption
scores. Each embodiment equally averages its fixed 100 episodes. Robot Avg
equally weights the two robot tracks, with no combined human--robot score.
Displayed subset scores use one decimal with half-up rounding. Robot Avg
is computed from the displayed subset percentages using the same rule.

\textbf{Judging.} A fixed GPT-6 Astra judge evaluates candidate captions
against reference requirements and sampled source evidence, with model
identities withheld. It uses high reasoning with a 12,000-token response
budget for human episodes and 20,000 for robot episodes. One truncated human
judgment receives a single 24,000-token recovery attempt with unchanged
inputs. The first valid judgment is retained. Because Astra is also an
evaluated system, generator--judge self-preference is a limitation.
The scores depend on the reference requirements and sampled visual evidence.
Camera timestamp agreement does not independently verify hardware
synchronization.

\subsection{Output validity and incomplete measurements}
\label{sec:annotation-failures}
Deterministic normalization may remove complete JSON fences, unwrap a single
unambiguous container, or discard unrelated metadata. It does not rewrite
captions or alter predicted intervals, apart from the stated endpoint rule.
Native outputs use the trained schema and production parser. Cache reuse
requires matching sources, intervals, prompts, model settings, and inputs.
Technical recovery preserves the request. Returned semantic answers and
refusals are not resampled to improve results.

\Needspace{3\baselineskip}
\textbf{Fixed-denominator reporting.} Failed generation and unavailable
judgments receive no operational credit. All episode, reference, and
prediction denominators remain fixed, and resolved contributions within
partially measured episodes are retained. Unavailable judgment is a
measurement status, not a finding that the caption is factually wrong.
We do not filter to the intersection of model-specific successful cases.
The reported \name WGO evaluation has complete oracle and predicted-caption
coverage. In the robot tracks of DenseCaptionBench, five sources lack new caption
generation and 98 judge jobs remain unavailable across oracle and joint
tasks. Previously measured temporal scores for the unchanged segmenter are
retained. These gaps affect the operational scores and preclude interpreting
ranking marks as a fully observed ordering.

\textbf{Technical recovery.} WGO baseline requests generally allow up to
three transport attempts with a 1,200-second timeout. Kimi and GPT-5.6 Luna
use first-attempt generation. Robot baseline generation allows at most two
attempts with a 300-second timeout. Archived baseline judgments allow up to
six cumulative technical attempts, with at most 900 seconds for the final
attempt. New \name RL judgments allow at most two identical attempts.
Recovery is restricted to technical failures with no retained valid verdict.
Ambiguous transmitted requests and refusals are not reissued under the
final RL protocol. These historical recovery budgets differ across runs
and do not constitute matched service-compute budgets.

\subsection{WGO-Bench protocol}
\label{sec:wgo-protocol}
WGO-Bench~\citep{macrodata2026wgo} contains 25 HomER, 50 DROID, and 25 Galaxea
episodes with 470, 150, and 123 reference segments, respectively.
The benchmark provides single-view videos and episode instructions.
Following its public evaluation protocol, scoring copies snap the first
predicted start and last predicted end to the corresponding reference
outer edges. Raw predictions remain unchanged. Segmentation uses
maximum-cardinality one-to-one matching at tIoU $\geq0.75$. Gold-window
caption accuracy uses all reference targets. Joint matching additionally
requires caption acceptance under the public text-only rubric, judged by
Gemini 3.5 Flash at temperature zero with an 8,192-token budget.
Counts are pooled within each subset, and reported averages equally weight
the three subset scores. The displayed average follows the rounded-subset
convention used in the main tables. Where intervals are reported, we use
2,000 paired episode-bootstrap draws stratified by subset.

All systems receive the episode instruction in both generation stages.
Reference caption text is never supplied. \name uses the single-view inputs
and native clocks with its fixed production workflow, without additional
benchmark-specific event merging or refinement. Thus the WGO results test
transfer to the benchmark's annotation policy without tuning boundaries to
its reference labels.

\subsection{Supplementary pipeline comparisons}
\label{sec:wgo-pipelines}

\begin{table}[!htb]
\centering
\floraTableStyle
\setlength{\tabcolsep}{0.45pt}
\hypersetup{citecolor=black!55}
\caption{\textbf{WGO-Bench pipeline comparison.} Subset scores and equal-weight averages (\%, $\uparrow$). Subsequent off-the-shelf model comparisons use the \hyperref[par:common-pipeline]{common pipeline} as a simple reference protocol that standardizes input construction and prompts while limiting API-call and visual-token overhead.}
\label{tab:wgo-pipelines-full}
\begin{tabular*}{\linewidth}{@{\extracolsep{\fill}}
  >{\raggedright\arraybackslash}m{0.248\linewidth}
  *{3}{M{0.067\linewidth}M{0.062\linewidth}M{0.071\linewidth}M{0.042\linewidth}}@{}}
\toprule
& \multicolumn{4}{c}{\shortstack{\textbf{Segmentation}\\F1@0.75}} &
  \multicolumn{4}{c}{\shortstack{\textbf{Captioning}\\Gold-window accuracy}} &
  \multicolumn{4}{c}{\shortstack{\textbf{Joint annotation}\\Semantic F1@0.75}} \\
\cmidrule(lr){2-5}\cmidrule(lr){6-9}\cmidrule(l){10-13}
\textbf{Pipeline / backbone} & {\fontsize{8}{9}\selectfont HomER} & DROID & {\fontsize{8}{9}\selectfont Galaxea} & \textbf{Avg} &
  {\fontsize{8}{9}\selectfont HomER} & DROID & {\fontsize{8}{9}\selectfont Galaxea} & \textbf{Avg} & {\fontsize{8}{9}\selectfont HomER} & DROID & {\fontsize{8}{9}\selectfont Galaxea} & \textbf{Avg} \\
\midrule
\addlinespace[2pt]
\multicolumn{13}{@{}l}{\textcolor{black!55}{\footnotesize Shared backbone: \textbf{Gemini 3.5 Flash}~\citep{google2026gemini35flash}}} \\
\addlinespace[3pt]
Common pipeline & 13.2 & 28.6 & 34.0 & 25.3 & 55.7 & 56.7 & 59.3 & 57.2 & 10.7 & 19.7 & 26.4 & 18.9 \\
\addlinespace[3pt]
Refiner~{\footnotesize\citeyearpar{macrodata2026refiner}} & 22.0 & 31.4 & 46.9 & 33.4 & 57.7 & 65.3 & 69.9 & 64.3 & 15.8 & 26.5 & 39.4 & 27.2 \\
\addlinespace[3pt]
LeRobot annotate~{\footnotesize\citeyearpar{lerobot2026v060}} & 14.2 & 28.3 & 31.0 & 24.5 & 47.9 & 34.7 & 41.5 & 41.4 & 8.0 & 20.3 & 22.9 & 17.1 \\
\midrule
\addlinespace[2pt]
\multicolumn{13}{@{}l}{\textcolor{black!55}{\footnotesize Model-specific backbones}} \\
\addlinespace[3pt]
Pegasus 1.5~{\footnotesize\citeyearpar{twelvelabs2026pegasus15}} & 10.7 & 44.8 & 28.1 & 27.9 & 48.1 & 52.7 & 49.6 & 50.1 & 8.8 & 34.0 & 27.1 & 23.3 \\
\addlinespace[3pt]

\mbox{\textbf{\name (ours)}}   & \floraOurScore{23.9} & \floraOurScore{56.1} & \floraOurScore{70.8} & \floraOurBest{50.3} & \floraOurScore{64.0} & \floraOurScore{55.3} & \floraOurScore{74.8} & \floraOurBest{64.7} & \floraOurScore{17.9} & \floraOurScore{32.5} & \floraOurScore{53.1} & \floraOurBest{34.5} \\
\bottomrule
\end{tabular*}
\end{table}

\textbf{Protocol.} The common pipeline, Refiner~\citep{macrodata2026refiner},
and LeRobot annotate~\citep{lerobot2026v060} share Gemini 3.5 Flash in
Figure~\ref{fig:wgo-pipelines}, following Refiner's published reference
setting~\citep{macrodata2026wgo}. Table~\ref{tab:wgo-pipelines-gemini38}
repeats the comparison with Gemini 3.8 Flash~\citep{google2026gemini38flash}.
These are our evaluations on the fixed WGO population. Both backbones use
temperature zero and a 32,768-token output limit. Refiner and LeRobot retain
their native prompts and visual plans at revisions \texttt{e402a667} and
\texttt{eb7a54c6}. Pegasus 1.5~\citep{twelvelabs2026pegasus15} and trained
\name retain their own backbones.

For oracle captioning, Refiner uses its fixed-window labeling prompt without
a seed caption, and LeRobot uses its visual-description component on each
reference interval. Pegasus receives an indexed list of reference intervals
with full-episode video context. Every system receives the episode instruction,
but no reference caption text. All three tasks use the same WGO scorer.
Native visual representations, prompt sequences, and compute remain different.

\textbf{Backbone sensitivity.} Refiner's mean joint F1 increases from
27.2\% with Gemini 3.5 Flash to 36.4\% with Gemini 3.8 Flash, compared with
34.5\% for trained \name. \name retains the highest mean segmentation
score, while Refiner leads caption accuracy and joint F1 in the Gemini 3.8
comparison. Pipeline rankings therefore depend on the generation backbone.

\textbf{API substitution.} As a diagnostic, we also replace both \name
models with the shared API backbone while retaining production windows,
prompts, continuation stitching, and caption context. Selected frames are
sent as timestamped PNG images rather than native video tokens. A bare
segment array may be wrapped in the required object without changing its
entries. With Gemini 3.5 Flash, this variant reaches 19.4\% segmentation F1,
52.1\% oracle accuracy, and 11.1\% joint F1. Gemini 3.8 Flash improves these
to 21.2\%, 72.1\%, and 16.4\%. Neither substitution matches the trained
system, so these experiments do not establish an advantage of the workflow
alone.

The Gemini 3.8 reruns preserve one source exclusion following a prior refusal,
including its 20 reference targets in the denominators. Previously completed
common-pipeline and Pegasus outputs are retained. Reruns use a 300-second
timeout and at most three explicit service-error attempts. Native format
correction and adaptive window processing remain part of their respective
pipelines. Unavailable captions receive no semantic match credit, while
valid segmentation outputs remain eligible for temporal scoring.

\begin{table}[!htbp]
\centering
\floraTableStyle
\caption{\textbf{Pipeline comparison with Gemini 3.8 Flash.} Scores are
percentages ($\uparrow$). Avg equally weights the three WGO subsets.
Green numbers identify \name variants. Shaded cells mark the best averages, including ties. The Pegasus
and trained \name results match Figure~\ref{fig:wgo-pipelines}.}
\label{tab:wgo-pipelines-gemini38}
\begin{tabular*}{\linewidth}{@{\extracolsep{\fill}}llcrrr@{}}
\toprule
\textbf{Pipeline} & \textbf{Backbone} & \textbf{Subset} &
\shortstack{\textbf{Segmentation}\\F1@0.75} &
\shortstack{\textbf{Captioning}\\Accuracy} &
\shortstack{\textbf{Joint}\\F1@0.75} \\
\midrule
\multirow{4}{*}{Common pipeline} & \multirow{4}{*}{\small Gemini 3.8 Flash} & HomER & 23.9 & 63.0 & 20.6 \\
 &  & DROID & 30.8 & 70.0 & 23.7 \\
 &  & Galaxea & 24.5 & 79.7 & 20.9 \\
 &  & \textbf{Avg} & 26.4 & 70.9 & 21.7 \\
\addlinespace[4pt]
\multirow{4}{*}{Refiner} & \multirow{4}{*}{\small Gemini 3.8 Flash} & HomER & 26.3 & 60.0 & 20.7 \\
 &  & DROID & 48.2 & 72.7 & 41.0 \\
 &  & Galaxea & 51.5 & 88.6 & 47.6 \\
 &  & \textbf{Avg} & 42.0 & \floraRankFirst{73.8} & \floraRankFirst{36.4} \\
\addlinespace[4pt]
\multirow{4}{*}{LeRobot annotate} & \multirow{4}{*}{\small Gemini 3.8 Flash} & HomER & 18.7 & 58.7 & 14.1 \\
 &  & DROID & 35.2 & 49.3 & 28.7 \\
 &  & Galaxea & 48.9 & 52.0 & 40.9 \\
 &  & \textbf{Avg} & 34.3 & 53.3 & 27.9 \\
\addlinespace[4pt]
\multirow{4}{*}{\name (API)} & \multirow{4}{*}{\small Gemini 3.8 Flash} & HomER & \floraOurScore{12.3} & \floraOurScore{56.6} & \floraOurScore{9.9} \\
 &  & DROID & \floraOurScore{18.1} & \floraOurScore{72.7} & \floraOurScore{13.2} \\
 &  & Galaxea & \floraOurScore{33.1} & \floraOurScore{87.0} & \floraOurScore{26.1} \\
 &  & \textbf{Avg} & \floraOurScore{21.2} & \floraOurScore{72.1} & \floraOurScore{16.4} \\
\addlinespace[4pt]
\multirow{4}{*}{Pegasus 1.5} & \multirow{4}{*}{\small Pegasus 1.5} & HomER & 10.7 & 48.1 & 8.8 \\
 &  & DROID & 44.8 & 52.7 & 34.0 \\
 &  & Galaxea & 28.1 & 49.6 & 27.1 \\
 &  & \textbf{Avg} & 27.9 & 50.1 & 23.3 \\
\addlinespace[4pt]
\multirow{4}{*}{\textbf{\name}} & \multirow{4}{*}{\small Qwen3.5 9B / 27B} & HomER & \floraOurScore{23.9} & \floraOurScore{64.0} & \floraOurScore{17.9} \\
 &  & DROID & \floraOurScore{56.1} & \floraOurScore{55.3} & \floraOurScore{32.5} \\
 &  & Galaxea & \floraOurScore{70.8} & \floraOurScore{74.8} & \floraOurScore{53.1} \\
 &  & \textbf{Avg} & \floraOurBest{50.3} & \floraOurScore{64.7} & \floraOurScore{34.5} \\
\bottomrule
\end{tabular*}
\end{table}
\FloatBarrier

\FloatBarrier
\section{Reward Evaluation Protocol}
\label{sec:reward-eval-protocol}

All reward results in Section~\ref{sec:reward-eval} use the same
\name-Rewarder model.

\paragraph{Baselines.}
General-purpose baselines receive the task instruction and the frames given
to \name-Rewarder, with the same prompts and answer formats. API models
request medium reasoning where supported, and reasoning compute is not
equated across models. Results for specialized reward models are taken from
their publications and keep their original protocols. RynnValue ranks 976
trajectories rather than 567, and the Robometer row takes the best value
reported across its variants.

\paragraph{RoboRewardBench.}
\name-Rewarder answers in score-only mode from
$\min(64,\max(2,\lfloor 10T\rfloor))$ uniformly sampled frames, with greedy
decoding and thinking disabled. MAE is computed on the 1--5 scale and
multiplied by 100. Baselines allow 8,192 output tokens, and only truncated
queries are retried with 16,384. Qwen3.5-397B-A17B receives 32 frames and the
other baselines up to 64. Qwen3.5-27B answers with thinking disabled. Kimi
K3's MAE covers 2,815 of the 2,831 episodes.

\paragraph{RobometerBench.}
Policy ranking uses 567 trajectories from six OOD datasets. Kendall's
$\tau_a$ is averaged over tasks within each dataset and then equally over
datasets. \name-Rewarder ranks trajectories by $\hat{y}$
(Eq.~\eqref{eq:outcome-expectation}), and general-purpose models by their
integer scores. VOC is the Pearson correlation between predicted progress and
elapsed time on 255 ID and 105 OOD successful trajectories. For
\name-Rewarder, let $q_t=\sum_{k}w_k\,p_\psi(a_t=k)$ be the expected quality
of step $t$ under the dense reward, with weights $w_k$ of $0$, $0.5$, and $1$
for failure, suboptimal, and normal. Progress at step $t$ is then
$\hat{y}\sum_{\tau\leq t}q_\tau/\sum_{\tau}q_\tau$. General-purpose models
predict integer progress scores. Progress that grows linearly with time scores
100 on this metric.

\paragraph{DenseRewardBench.}
The 40 episodes are excluded from dense training
(Appendix~\ref{sec:reward-dense-details}). Labels are compared at 1\,Hz on
4,861 time steps (3,942 normal, 619 suboptimal, and 300 failure), and all
metrics pool these steps. All models label the same 508 windows of 64 frames
at 4\,FPS, and predictions are stitched by Eq.~\eqref{eq:dense-stitching},
dropping the last 4\,s of every non-final window. \name-Rewarder decodes
greedily under its answer format with thinking disabled. Baselines receive
each window as 64 still images (32 for Qwen3.5-397B-A17B), with the same
prompt and a note on frame grouping. They answer in free text with 8,192
output tokens, and only truncated malformed answers are retried with 16,384
and then 32,768 tokens.

\paragraph{RubricBench.}
The 40 teleoperated episodes come from 10 tasks and share no episode or task
name with the rubric training pool. Each camera is a separate video, sampled
at 1\,FPS with at most 83 frames for baselines and at 2\,FPS with at most 166
frames for \name-Rewarder. \name-Rewarder answers in thinking mode with
temperature $1.0$ and up to 20,480 response tokens. It draws nine samples per
episode and answers with their medoid, the sample with the smallest summed L1
distance to the others. We report the expected error of this answer over
random draws of nine from 32 samples per episode. Baselines use one sample
with the same decoding settings, with two exceptions. GPT-6 Astra and Qwen3.5-27B were
evaluated with greedy decoding and a 16,384-token budget. The GPT-5.6 models
and Claude Sonnet 5 do not apply a temperature setting. GPT-6 Astra's parse
failure is an unserved request. On the 39 answered episodes its L1 is 3.7.
Kimi K3's parse failure exhausted the response budget.

\FloatBarrier
\section{Annotation Cost and Throughput}
\label{sec:annotation-cost-estimation}

\subsection{Evaluation cohort and accounting}
\begin{figure}[!htbp]
\centering
\captionsetup{justification=raggedright,singlelinecheck=false,font=small,skip=6pt}
\includegraphics[width=\linewidth]{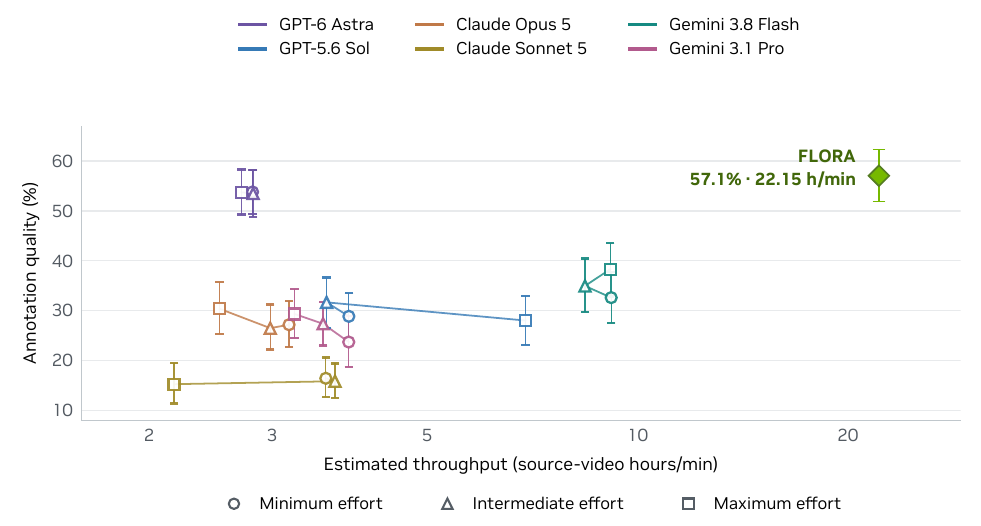}
\caption{\textbf{Annotation quality versus conditional throughput.}
The 50-episode quality scores match Figure~\ref{fig:annotation-cost}.
API rates follow the quota scenarios in
Table~\ref{tab:annotation-quota-scenarios}. The \name rate assumes 12
independent eight-GPU replicas. Vertical bars are episode-bootstrap 95\%
intervals. Throughput is modeled, not measured, and the points are not
budget matched.}
\label{fig:annotation-throughput}
\end{figure}

Figures~\ref{fig:annotation-cost} and \ref{fig:annotation-throughput} use
50 fixed DenseCaptionBench episodes: 20 human, 15 parallel-gripper, and 15 dexterous-hand
episodes, totaling 4,483.54 seconds of source video. Quality equally averages
segmentation F1@0.5, oracle caption point quality, and joint point F1@0.5
within each embodiment, then equally averages the three embodiments.
The subset differs from the complete DenseCaptionBench evaluation in
Table~\ref{tab:internal-e2e}.

\textbf{Quality uncertainty.} Failed or unavailable contributions receive
no operational credit under Appendix~\ref{sec:annotation-failures}.
\name uses the selected RL pair and its current reference snapshot.
API results retain the cost study's original reference snapshot.
The resulting comparison is not fully reference-version matched.
Figure~\ref{fig:annotation-throughput} shows 95\% intervals from 10,000
paired episode-bootstrap draws within embodiment. Figure~\ref{fig:annotation-cost}
shows mean estimates only. The intervals quantify episode-sampling uncertainty,
excluding generation, judge, reference, and service uncertainty.
Unresolved native contributions separately bound mean quality at
57.1--59.6\%. The displayed points and retained bootstrap intervals use operational
scores with unavailable contributions assigned no credit.

\textbf{Reasoning settings.} Minimum, intermediate, and maximum denote
requested provider settings. GPT-6 Astra, GPT-5.6 Sol, Claude Opus 5, and
Claude Sonnet 5 use \texttt{low}, \texttt{medium}, and \texttt{max}.
Gemini 3.8 Flash and Gemini 3.1 Pro use \texttt{low}, \texttt{medium}, and
\texttt{high}. These settings are not compute matched across providers.
The left panel of Figure~\ref{fig:annotation-cost} compares \name with
maximum-effort Astra, Claude Opus 5, and Gemini 3.8 Flash. Its dashed line
traces the empirical API-only Pareto frontier. Connecting lines are visual
guides, not interpolated evaluations.
Pareto comparisons concern displayed mean estimates under the stated
assumptions, not statistical dominance.

\textbf{API cost.} We price returned usage at public standard rates from
October 4, 2026 and divide by source-video hours. The accounting includes
segmentation and predicted-interval caption generation, excluding training,
reference construction, oracle captioning, and judging. Input, cache-read,
cache-write, and output usage are counted once, with reasoning included in
output tokens. All physical attempts with returned usage are included.
Of 7,062 attempts, 6,956 report usage and 106 remain unpriced. Affected
costs are therefore known-usage subtotals, not complete invoices.

\subsection{Native deployment model}
The \name cost estimate uses the same model sizes as the evaluated system,
a 9B segmenter and a 27B captioner. Its reference workload of 660 windows
and 890 captions comes from the earlier deployment study. It is a workload
assumption, not a measured invocation count or serving rate for the RL pair.
An eight-GB300 deployment model combines peak compute, bandwidth, and a
production-rate envelope, assuming data parallelism and batch size 16.
The conditional rate is $S=1.23495$ episodes/s, with engineering scenarios
of $0.70804$--$1.74862$ episodes/s. The production GPU-hour denominator
has not been verified.

For $N$ episodes, $g$ GPUs, total source duration $H$ hours, and an assumed
GPU-hour price $p$, the cost per source-video hour is
\begin{equation}
  C_{\mathrm{Flora}}=\frac{pgN}{3600HS}.
  \label{eq:annotation-cost}
\end{equation}
With $N=50$, $H=1.24543$, $g=8$, and $p=10$ USD/GPU-hour, this gives
\$0.7224 per source-video hour, with scenarios of \$0.5102--1.2600.
Against maximum-effort Astra (\$88.4922), the cost ratio is $122.49$,
summarized as approximately $125\times$.
These ranges represent engineering scenarios, not confidence intervals.
Figure~\ref{fig:annotation-cost} shows the reference cost only. GPU pricing is assumed.
Storage, billing granularity, and deployment overheads are excluded.

\subsection{Conditional throughput comparison}
\label{sec:annotation-throughput-details}
For API setting $m$, let $A_m$ be physical segmentation and predicted-caption
requests and $T_{m,k}$ the accounted token demand of quota type $k$ over
the cohort. A request-per-minute limit $R_m$ and token-per-minute limits
$L_{m,k}$ give the idealized quota ceiling
\begin{equation}
  Q_m=H\min\left(\frac{R_m}{A_m},
    \min_{k\in\mathcal{K}_m}\frac{L_{m,k}}{T_{m,k}}\right)
  \quad\text{source-video hours/min}.
  \label{eq:annotation-quota-throughput}
\end{equation}
Table~\ref{tab:annotation-quota-scenarios} lists the selected quota scenarios.
They are fixed comparison assumptions from the October 2026 analysis,
not verified account entitlements or universal provider limits.

\begin{table}[!htbp]
\centering
\floraTableStyle
\caption{\textbf{Quota scenarios for throughput estimates.} RPM denotes
requests per minute. M denotes million tokens per minute.}
\label{tab:annotation-quota-scenarios}
\begin{tabularx}{\linewidth}{@{}lrrX@{}}
\toprule
\textbf{Models} & \textbf{RPM} & \textbf{Input/admission} & \textbf{Output} \\
\midrule
GPT-6 Astra / GPT-5.6 Sol & 15,000 & 40M & Included in admission proxy \\
Claude Opus 5 / Sonnet 5 & 10,000 & 10M & 2M, including reasoning \\
Gemini 3.8 Flash & 20,000 & 20M & No separate allowance modeled \\
Gemini 3.1 Pro & 2,000 & 8M & No separate allowance modeled \\
\bottomrule
\end{tabularx}
\end{table}

The OpenAI scenario uses $\max(32{,}768,\text{prompt tokens})$ per attempt
as an admission proxy. Actual multimodal quota debits are unverified.
Claude input demand includes uncached input and cache writes, excluding
cache reads. Gemini demand includes cached input in its input-token quota.
The selected scenarios follow provider rate-limit documentation and the
comparison profiles available at analysis time.\footnote{%
\href{https://developers.openai.com/api/docs/guides/rate-limits}{OpenAI rate limits},
\href{https://platform.claude.com/docs/en/api/rate-limits}{Claude rate limits}, and
\href{https://ai.google.dev/gemini-api/docs/rate-limits}{Gemini rate limits}.}
Missing usage makes some token demands incomplete. Attained service speed,
spending limits, daily/monthly quotas, and dedicated provisioning are not
modeled. The estimates assume sufficient concurrency and stable service.

For $r$ independent eight-GPU replicas, native throughput is modeled as
\begin{equation}
  Q_{8r}=r\frac{60HS}{N}.
  \label{eq:annotation-native-throughput}
\end{equation}
One replica gives $1.8457$ source-video hours/min. We display 12 replicas
(96 GPUs), yielding $22.15$ hours/min at an assumed \$16.00/min.
This configuration fits within the lowest quota-implied API spending rate
in the comparison (\$16.08/min). API points retain their individual quota
profiles and spending rates, so the comparison is not pointwise budget
matched. Linear replica scaling is assumed, with no measured 96-GPU
serving result or model for storage, decoding, networking, and interference.

\FloatBarrier

\FloatBarrier
\clearpage
\section{Additional Qualitative Evidence}
\label{sec:additional-qualitative}

\begin{figure}[H]
  \centering
  \includegraphics[width=\linewidth]{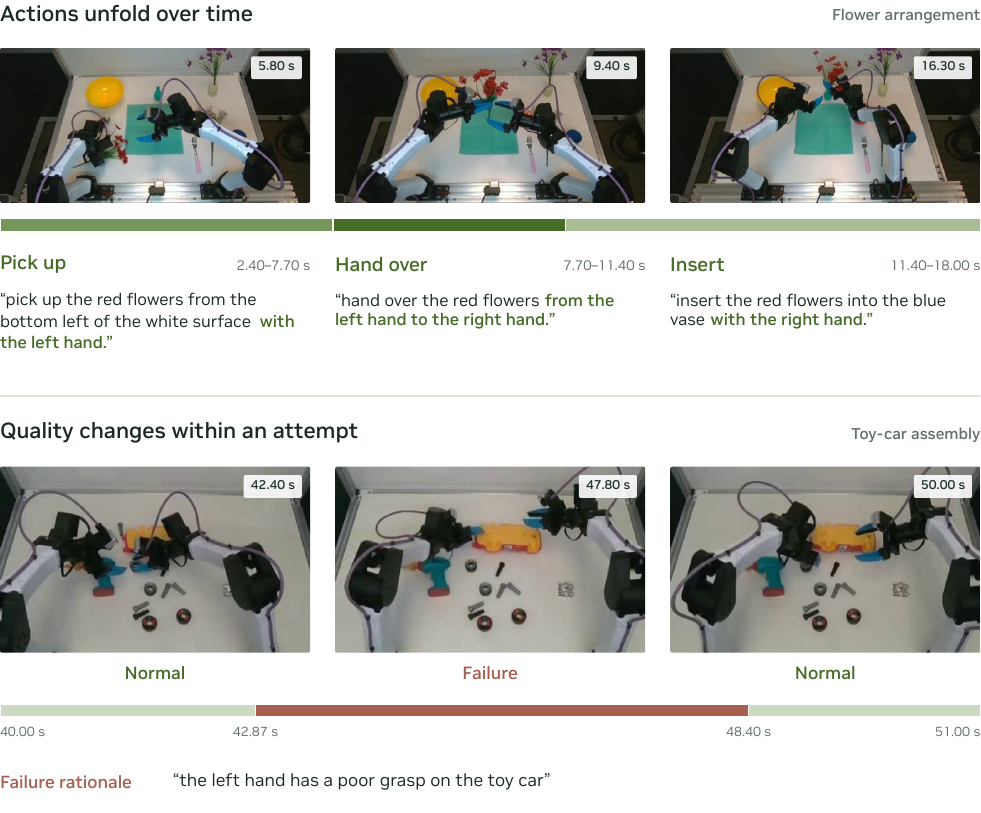}
  \caption{\textbf{Actions and rewards over time.} Three consecutive actions (top) and a localized grasp failure (bottom), from two separate episodes.}
  \label{fig:episode-sequence}
\end{figure}

\begin{figure}[H]
\centering
\includegraphics[width=\linewidth]{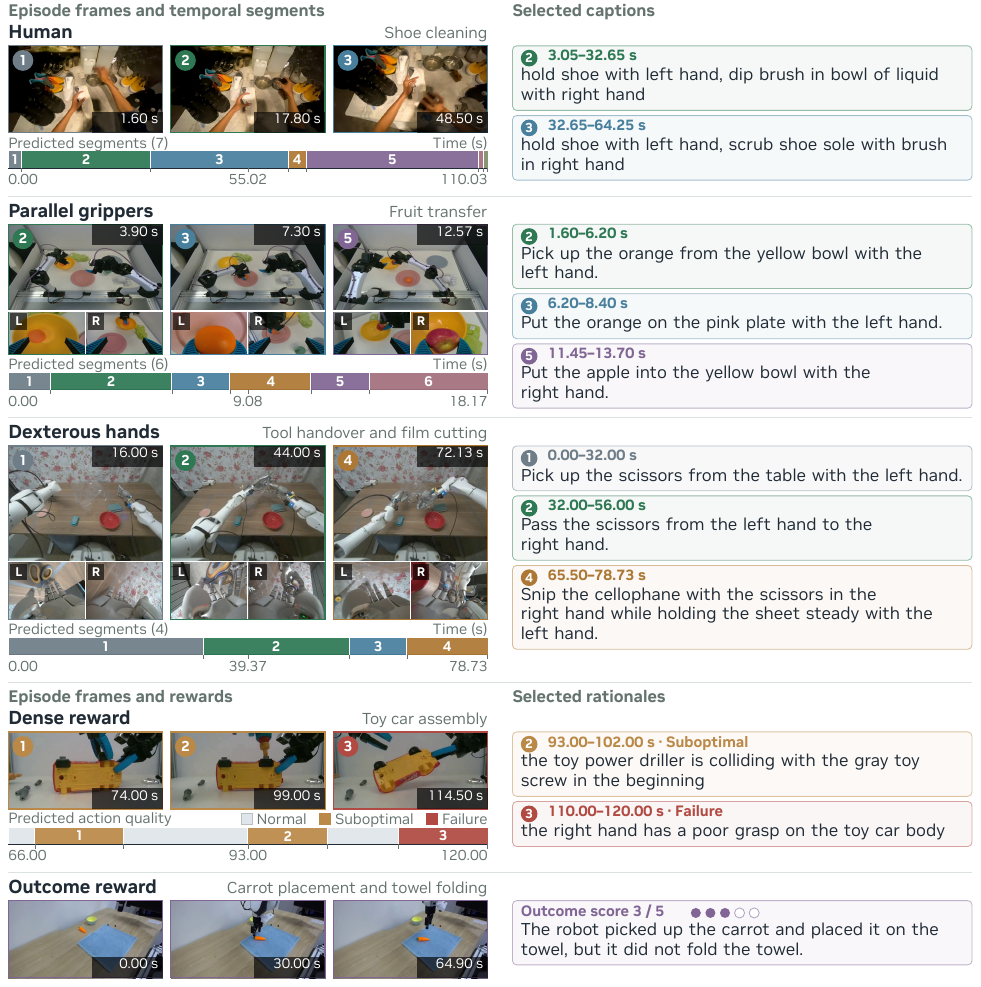}
\caption{\textbf{Example episode outputs across embodiments.} Timestamped intended-action captions appear above dense execution-quality and outcome labels. Robot mosaics retain synchronized scene and wrist views. Selected captions and rationales illustrate the structured records generated by the engine.}
\label{fig:episode-annotation-demo}
\end{figure}
\begin{figure}[H]
  \centering
  \includegraphics[width=0.90\linewidth]{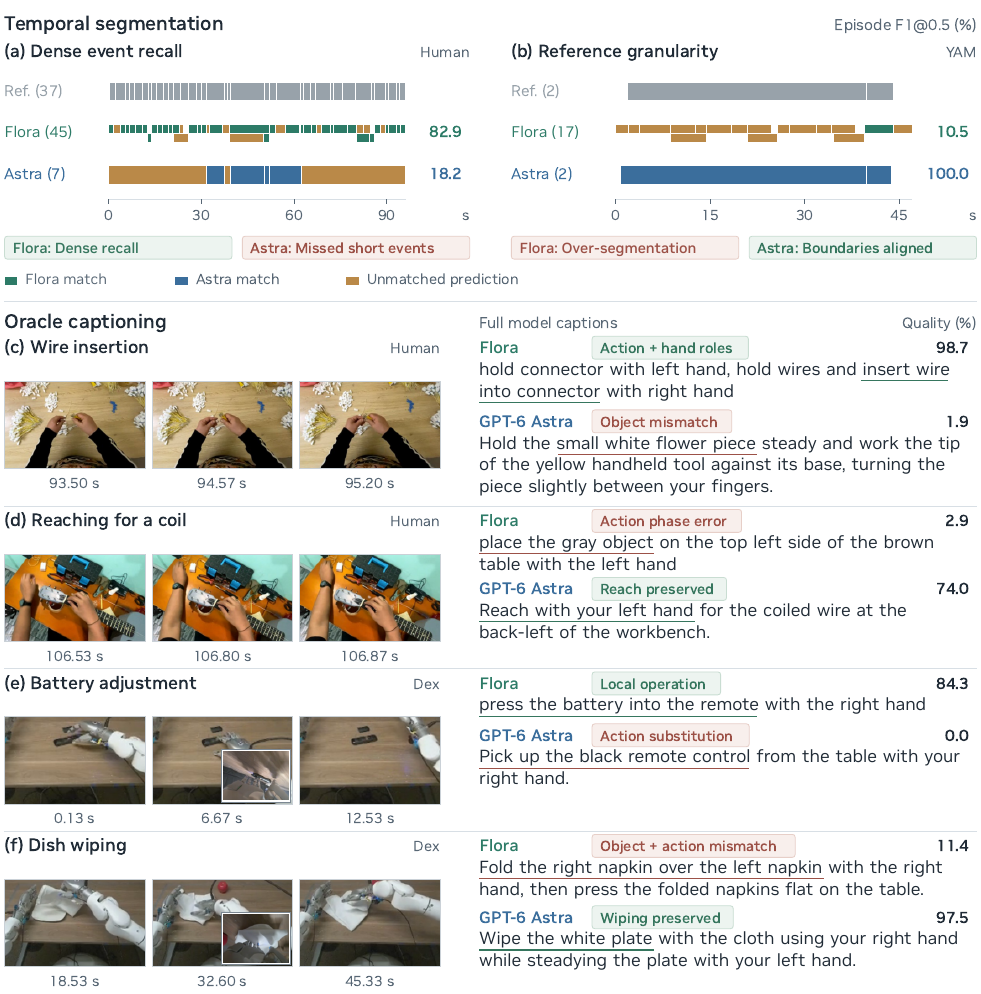}
  \captionsetup{justification=raggedright,singlelinecheck=false,font=small}
  \caption{\textbf{Qualitative comparison with GPT-6 Astra.} \name captures better dense actions (a) and local manipulation details (c,e), but can over-segment coarse events (b) and confuse action phases or objects (d,f). Parentheses denote segment counts. Temporal and caption scores are F1@0.5 and point quality on reference intervals, respectively. Robot insets show right-wrist views.}
  \label{fig:annotation-qualitative-full}
\end{figure}

We qualitatively compare \name to GPT-6 Astra on captioning and reward labeling. For annotation, we use paired episode diagnostics and a reference-informed sampled review of 12 oracle windows. Figure~\ref{fig:annotation-qualitative-full} shows two temporal examples and four caption examples using fixed references and fixed-protocol judgments.

\clearpage

\textbf{Temporal coverage and granularity.} Panel~(a) illustrates the advantage on dense actions: \name matches nearly all reference intervals, whereas Astra matches only a few. The additional segments can also fragment coarse reference events: in panel~(b), \name splits two references into many short segments and matches only one, while Astra matches both. On mutually observed dexterous-hand episodes, recall is similar, but \name is less precise. The aggregate parallel-gripper ordering reverses on mutually observed episodes.

\textbf{Local operations and object identity.} \name preserves wire insertion in panel~(c), where Astra misidentifies a connector as a flower piece, and battery adjustment in panel~(e), where Astra substitutes remote pickup. Conversely, \name substitutes placement for reaching in panel~(d) and napkin folding for dish wiping in panel~(f). On jointly judged human windows, \name covers coordination requirements far more often than Astra (58.3\% versus 21.2\%), while action coverage is similar. Different input pipelines and reference types limit causal attribution.

\clearpage
\begin{figure}[H]
  \centering
  \includegraphics[width=0.90\linewidth]{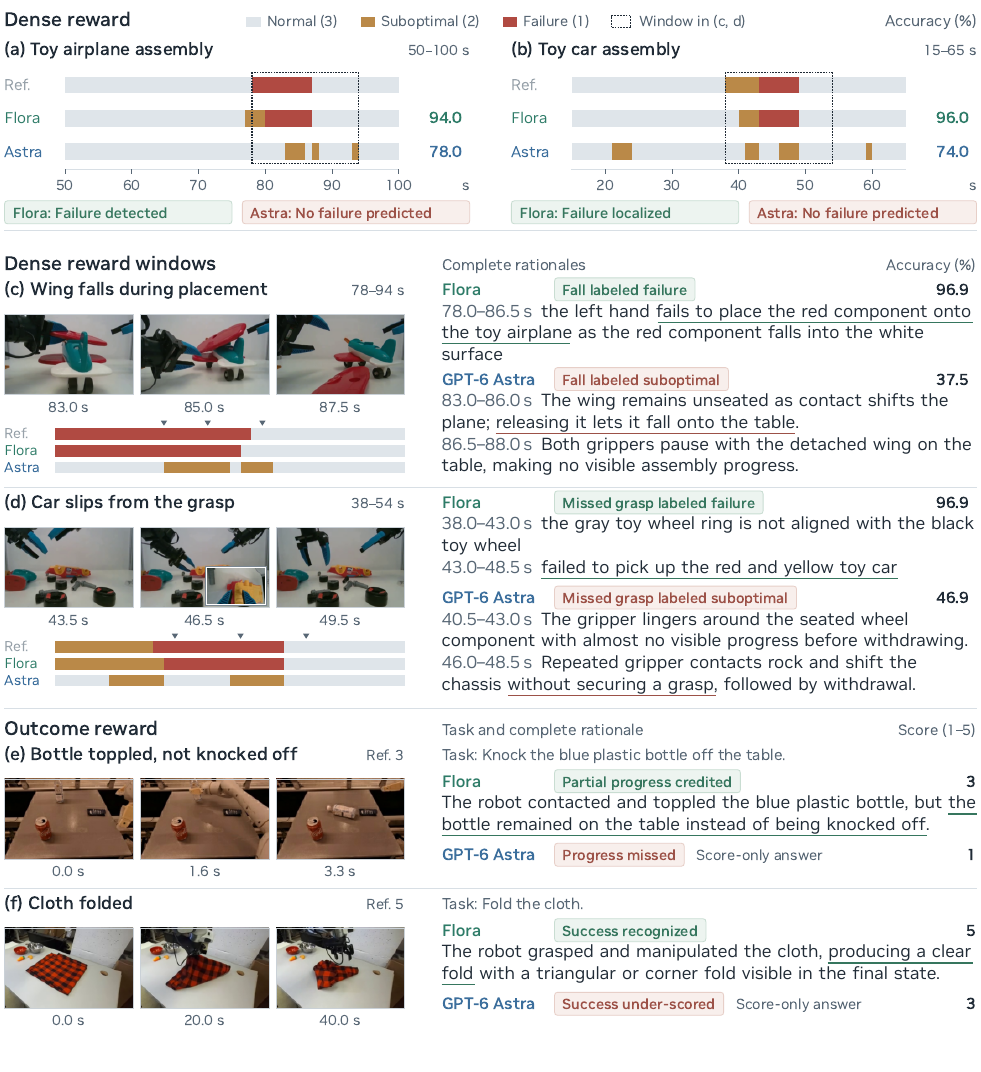}
  \captionsetup{justification=raggedright,singlelinecheck=false,font=small}
  \caption{\textbf{\name-Rewarder flags failures and credits progress that GPT-6 Astra misjudges.} Selected examples. (a--d)~DenseRewardBench: Astra calls a falling wing and a missed grasp suboptimal, although its own rationales describe both. (e,f)~RoboRewardBench: Astra scores a partial and a full success too low. (a,b)~50\,s excerpts with per-second labels. (c,d)~the dashed 16\,s windows. Right-hand numbers give agreement with the reference (\%).}
  \label{fig:reward-qualitative-full}
\end{figure}

\textbf{Failure severity.} Astra understates severity, even when its own rationale describes a failure. In panel~(c), it states that the wing falls onto the table but labels the drop suboptimal, and in panel~(d), it labels gripper contacts ``without securing a grasp'' as suboptimal. \name-Rewarder labels both events as failures. Over the longer excerpts in panels~(a) and~(b), its labels follow the reference segments. Higher failure recall has a cost: \name-Rewarder also labels about 11\% of normal steps as failures.

\textbf{Partial outcomes.} \name-Rewarder's advantage comes from partially completed tasks. On complete failures and successes, Astra is at least as accurate. In panels~(e) and~(f), Astra gives the lowest score to a bottle that was toppled but not knocked off the table, and scores a completed fold 3 instead of 5. \name-Rewarder's rationales name the progress behind each score.

\FloatBarrier
\clearpage
\section{Timestamped Captioning: Implementation and Training}
\label{sec:annotation-implementation}

\subsection{Temporal reconstruction and visual context}
The segmenter uses overlapping windows with a central emission core and
context guards (Section~\ref{sec:textual-annotation}). Window-local timestamps
are projected into the episode clock. Stitching associates compatible
observations using temporal overlap and continuation flags, while preserving
independent concurrent actions. Guard-only fragments do not create an event
for that window. A bounded set of shifted windows refines unresolved
boundaries. Unobserved endpoints retain the observed extent without
inventing a transition.

\textbf{Window-local predictions.} For sampled frames $X_{W_k}$ and a window prompt $q_k$, the segmenter predicts local intervals and continuation flags $\ell_{kj},r_{kj}$:
\begin{equation}
    \mathcal{F}_k=f_\theta(X_{W_k},q_k)
    =\{(u_{kj},v_{kj},\ell_{kj},r_{kj})\}_{j=1}^{n_k},
    \label{eq:temporal-fragments}
\end{equation}
The projection $\Phi_k$ converts local timestamps to episode time. Stitching $\mathcal{S}$ joins compatible fragments while preserving concurrent actions:
\begin{equation}
    \mathcal{I}=\{I_i\}_{i=1}^{N}
    =\mathcal{S}\!\left(\{(W_k,\Phi_k(\mathcal{F}_k))\}_{k=1}^{K}\right).
    \label{eq:episode-stitching}
\end{equation}

\textbf{Caption context.} After stitching, the nearest non-overlapping events
provide context for a target interval $I_i=[s_i,e_i)$:
\begin{equation}
    j_i^- \in \operatorname*{arg\,max}_{j:\,e_j\leq s_i} e_j,
    \qquad
    j_i^+ \in \operatorname*{arg\,min}_{j:\,s_j\geq e_i} s_j.
    \label{eq:caption-context}
\end{equation}
The corresponding visual blocks in Eq.~\eqref{eq:segment-caption} are
$X_i^-=X_{I_{j_i^-}}$, $X_i=X_{I_i}$, and $X_i^+=X_{I_{j_i^+}}$.
Ties favor the shorter neighboring interval. Unavailable neighbors are
omitted. Each block has its own frame budget and
retains source timestamps. These adjacent-event blocks are used for training
and production captioning. Evaluation on DenseCaptionBench instead restricts context
to at most 2 seconds on either side of the target, as specified in
Appendix~\ref{sec:annotation-inputs}.

\textbf{Multiple views.} At each sampled source time, the main camera occupies
the upper panel of a mosaic, with the left and right wrist cameras below.
This arrangement is fixed across time and across the before, target, and
after blocks. The prompt identifies the camera roles and requests one
account of the intended action across views. Single-view inputs preserve
the same temporal block structure.

\subsection{Supervised fine-tuning}
We initialize the segmenter $f_\theta$ and captioner $g_\phi$ from
Qwen3.5-9B and Qwen3.5-27B~\citep{qwen3.5}, respectively. Supervision comprises
timestamped action annotations from heterogeneous human and robot
manipulation recordings. Segmentation targets serialize local intervals and
continuation flags. Captioning targets pair an action instruction with its
annotated interval and visual context. Both models minimize cross-entropy
on output tokens only:
\begin{equation}
    \mathcal{L}_{\eta}(\mathcal{B}) =
    -\frac{\displaystyle\sum_{(x,y)\in\mathcal{B}}
        \sum_{t=1}^{|y|}\log p_{\eta}(y_t\mid x,y_{<t})}
        {\displaystyle\sum_{(x,y)\in\mathcal{B}}|y|},
    \qquad \eta\in\{\theta,\phi,\psi\}.
    \label{eq:annotation-training}
\end{equation}
Here $x$ contains the visual evidence and task prompt, and $y$ is the
segmentation output or action instruction. The segmenter updates rank-32
LoRA adapters with scaling $\alpha=64$ in the language model's attention and
MLP projections. The captioner updates the language model and visual merger,
while freezing the visual patch embedding and transformer blocks.
SFT uses AdamW with gradient clipping. Training uses sharding and activation checkpointing for the captioner.

\Needspace{9\baselineskip}
\subsection{Temporal segmentation reinforcement learning}
\label{sec:tempseg-rl}
We optimize the segmenter with GRPO~\citep{shao2024deepseekmath} using a
verifiable reward. Predictions and references are clipped to the emission
core, and guard-only fragments are discarded. Maximum-cardinality
one-to-one matching at temporal-IoU threshold $\tau$ defines F1 $F_\tau$ and
precision $P_\tau$. The reward is
\begin{equation}
    r_{\mathrm{seg}}=0.45F_{0.5}+0.15F_{0.75}+0.25P_{0.5}+0.15B.
    \label{eq:tempseg-rl-reward}
\end{equation}
For each matched pair at tIoU $0.5$, let $d$ be the sum of absolute endpoint
errors, augmented by 0.5 seconds for each incorrect continuation flag.
The boundary term $B$ averages $\max(0,1-d/(1\,\mathrm{s}))$ over these
pairs. Matching first maximizes cardinality, then minimizes total clamped
error. If no pair matches, $B=0$, except when both sets are empty, for
which $B=1$. Both-empty sets receive reward one. Invalid or truncated outputs
receive zero.

Rollout advantages subtract the group mean without standard-deviation normalization. Only LoRA adapters are updated, with KL regularization to the fixed SFT policy.

\textbf{Checkpoint selection.} A separate validation set determines the
reported checkpoint. Selection maximizes a prespecified weighted F1@0.5,
subject to at most a two-percentage-point drop in each guarded validation
group. Benchmark results are not used for selection.

\subsection{Captioner reinforcement learning}
\label{sec:captioner-rl}
The captioner uses a reference-grounded reward based on intended action,
object identity, manipulator coordination, and relevant detail. Source
video and reference requirements define supported mass $R$, covered mass
$M$, and factual-error penalty $F$:
\begin{equation}
    r_{\mathrm{cap}}=\frac{2M}{R+M+F}.
    \label{eq:caption-rl-reward}
\end{equation}
The group weights and phase gates follow the point-quality formulation in
Appendix~\ref{sec:annotation-metrics}. Reference instruction text is never
part of the captioner's generation input.

Rewards are centered within each rollout group without standard-deviation normalization. The language model is updated while the visual encoder and merger remain frozen. Reference instructions remain unavailable to the generation input.

\textbf{Checkpoint selection.} Validation equally weights the three
embodiments and precedes benchmark evaluation. We maximize mean quality
among checkpoints without per-track regression. If none qualifies, we
maximize the worst-track change. The segmenter and captioner are selected
independently.

\Needspace{20\baselineskip}
\subsection{Pipeline schematics}
\begin{figure}[H]
\centering
\includegraphics[width=\linewidth]{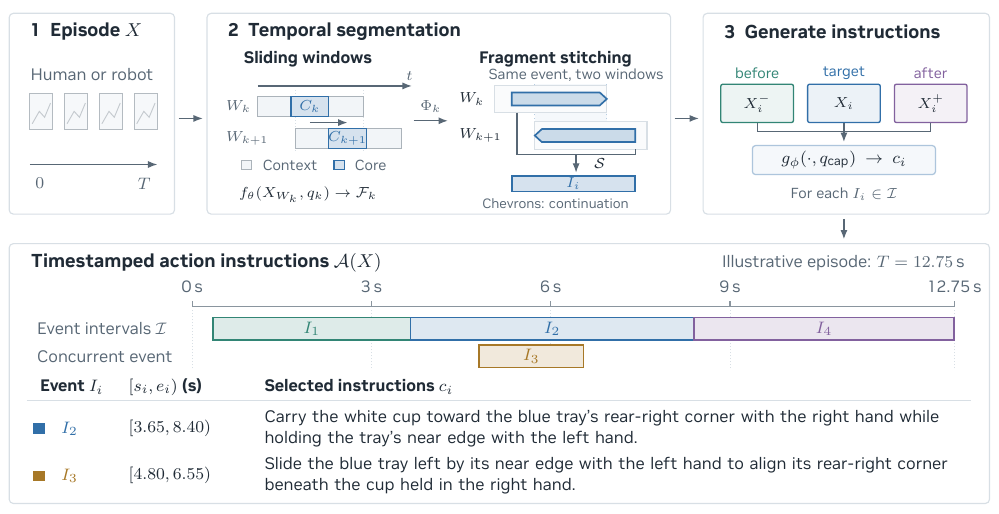}
\caption{\textbf{Windowed localization and contextual captioning.} Overlapping windows emit fragments from their central regions. Timestamp projection and stitching reconstruct episode intervals, which are captioned independently with before, target, and after visual evidence.}
\label{fig:textual-annotation}
\end{figure}
\begin{figure}[H]
\centering
\includegraphics[width=\linewidth]{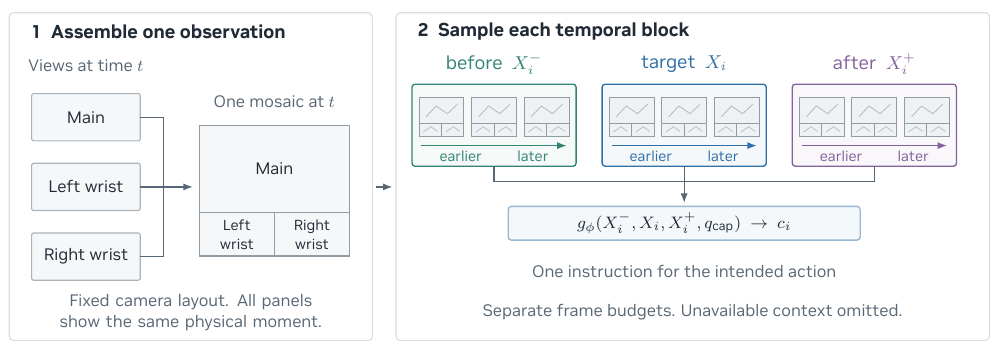}
\caption{\textbf{Synchronized views and temporal context.} Each mosaic aligns cameras at a source time. Mosaics retain their temporal order within the before, target, and after blocks.}
\label{fig:caption-input}
\end{figure}

\FloatBarrier
\clearpage
\section{Reward Modeling Implementation and Training}
\label{sec:reward-implementation}

This appendix details the inputs, data, prompts, and training of
\name-Rewarder (Section~\ref{sec:reward-modeling}). Benchmark protocols and
results appear in Section~\ref{sec:reward-eval}.

\textbf{Reward interfaces.} With a task instruction $\ell$, the outcome and dense interfaces return
\begin{equation}
    \mathcal{R}(X)=\big(\{a_t\}_{t},\,s\big),
    \qquad a_t\in\{1,2,3\},\quad s\in\{1,\dots,5\},
    \label{eq:reward-output}
\end{equation}
Here $a_t$ is execution quality for a 0.5-second group, and $s$ is the episode outcome. Score-token probabilities define the continuous outcome reward:
\begin{equation}
    p_k=p_\psi(s=k\mid X,\ell,q_{\mathrm{out}}),\qquad
    \mathbb{E}[s]=\sum_{k=1}^{5}k\,p_k,\qquad
    \hat{y}=\frac{\mathbb{E}[s]-1}{4}\in[0,1].
    \label{eq:outcome-expectation}
\end{equation}
For each dense window, the output contains its worst label, ordered quality sequence, and rationales for non-normal spans:
\begin{equation}
    r_\psi(X_{W_k},\ell,q_{\mathrm{dense}})
    =\Big(w_k,\;(a_{k,1},\dots,a_{k,32}),\;
      \{(u_{kj},v_{kj},c_{kj})\}_{j}\Big),
    \qquad w_k=\min_{g}\,a_{k,g},
    \label{eq:dense-window}
\end{equation}
Given a rubric $\rho$, grading is constrained to allowed values and compared with human grades by L1 error:
\begin{equation}
\hat{\mathbf{y}}=r_\psi(X,\ell,\rho,q_{\mathrm{rub}})
    \in V_1\times\cdots\times V_n,
    \qquad
    R_\rho(X)=\sum_{j=1}^{n}\hat{y}_j,
    \qquad
l(\hat{\mathbf{y}},\mathbf{y})=\sum_{j=1}^{n}\lvert\hat{y}_j-y_j\rvert,
    \label{eq:rubric-reward}
\end{equation}

\newtcblisting{rewardprompt}[1]{%
  listing only, breakable, enhanced,
  colback=black!2, colframe=black!30, boxrule=0.4pt, arc=1pt,
  left=4pt, right=4pt, top=2pt, bottom=2pt,
  title={#1}, fonttitle=\footnotesize\bfseries, coltitle=black,
  colbacktitle=black!7,
  listing options={basicstyle=\ttfamily\scriptsize, breaklines=true,
    breakindent=0pt, columns=fullflexible, keepspaces=true,
    showstringspaces=false, aboveskip=0pt, belowskip=0pt}}

\subsection{Video Input}
\label{sec:reward-video-input}

\textbf{Multi-view mosaics.}
Outcome and dense prompts receive one video in which each sampled time is a
single mosaic of up to four synchronized views. The layout is
$1\times1$, $1\times2$, $1\times3$, or $2\times2$ for one to four views.
Camera roles determine the placement. A $1\times3$ mosaic orders the left,
center, and right views. A $2\times2$ mosaic places the scene (top or front)
view at the upper left, a second scene or wrist view at the upper right,
and the fixed left and right cameras in the lower row. When more than
four views exist, scene and fixed side cameras take precedence over wrist
views, and downscaled duplicates of a stream are dropped. Each view is
resized into its cell without cropping or padding. Cell sides are multiples
of 32 pixels and target about $10^5$ pixels per view, and the row height is
chosen to minimize aspect-ratio distortion. Mosaics carry no gutters, borders,
or text. A
64-frame clip is capped at $14{,}336$ visual tokens. The bimanual
recordings used for dense rewards produce a $1152\times288$ mosaic for three views
(left, top, right) and a $768\times576$ mosaic for four views (top, bottom,
left, right), each with $384\times288$ cells. A 64-frame clip of the latter
occupies $13{,}824$ visual tokens.

\textbf{Timestamps.}
For outcome and dense prompts, we pass the source frame indices and frame
rate of every sampled frame to the model's video processor instead of
resampling inside it. The timestamp tokens of Qwen3.5 therefore refer to
absolute episode seconds in both training and inference, for every source
frame rate.

\textbf{Rubric inputs.}
Rubric prompts instead receive each camera as a separate video, labeled
\texttt{Video~1}, \texttt{Video~2}, and so on in the text. Frames are
sampled at 1\,FPS, uniformly subsampled to at most 82 frames per camera,
and resized to between $65{,}536$ and $131{,}072$ pixels each.

\subsection{Dense Reward}
\label{sec:reward-dense-details}

\textbf{Annotation protocol.}
Annotators fill one table per episode. Each row names an action, gives its
start and end times in seconds, and assigns an outcome of success, normal,
suboptimal, or failure. Success and normal both map to the normal label. A
free-text description of the visible issue is required for suboptimal and
failure actions. We convert the table to frame labels on the native 30\,FPS
clock. Every frame starts as normal, each action assigns its label to its
frames, and overlapping actions resolve to the more severe label
(failure $>$ suboptimal $>$ normal). The result is stored as contiguous
run-length segments with a label and issue text. A new segment begins
whenever either changes. Normal segments carry no text.

\textbf{Supervision.}
Dense labels come from human-annotated bimanual robot episodes with synchronized cameras. Training covers normal execution, recoverable inefficiencies, and failures. Episodes used for evaluation are excluded from training and validation.

\textbf{Window geometry.}
A window is defined by an anchor time $\alpha_k$. Its 64 positions sample the
source video at $\alpha_k+(i-9)/4$\,s for $i=1,\dots,64$, so the window spans
$[\alpha_k-2,\,\alpha_k+14)$\,s. Positions outside the episode repeat the
nearest frame. Group $g$ contains positions $2g-1$ and $2g$ and is valid if
either of its frames lies inside the episode. Inference anchors are
$\alpha_k=10k$\,s while $\alpha_k$ is below
$\alpha_{\max}=t_{\mathrm{last}}-9.75$\,s, followed by a final window at
$\alpha_{\max}$, where $t_{\mathrm{last}}$ is the time of the last frame.
The last frame of the episode thus falls at position 48, the end of group 24,
and the final window's last 4\,s repeat it. Every window has the same
geometry and output length. Each group takes the more severe label of its two frames. Padded groups are supervised with the label of the
boundary frame.

\textbf{Training examples.}
Windows are sampled to cover all three quality categories and temporal transitions. The split is made at the episode level. Non-normal labels include short evidence-grounded rationales from the human annotations.

\textbf{Task conditioning.}
Every dense prompt conditions on the episode's global task instruction,
never on the sub-task annotations. Words that reveal the outcome, such as
\emph{failure}, \emph{success}, \emph{retry}, and \emph{recovery}, are
removed. Placeholders and opaque assembly codes make an episode ineligible.
Task descriptions are normalized to identify the requested operation without revealing its outcome.

\textbf{Prompt and answer.}
The system prompt defines the score scale and the answer grammar:
\begin{rewardprompt}{Dense reward: system prompt}
You are a dense action-quality reward model for robot manipulation.
You watch a synchronized multiview clip and predict action quality at the finest temporal resolution visible to the model.

The clip contains exactly 64 sampled frames at 4 FPS. The vision encoder combines each consecutive pair of frames into one temporal group, so you must predict exactly 32 ordered groups. Group 1 contains frames 1-2, group 2 contains frames 3-4, and group 32 contains frames 63-64.
The clip is retrospective: a later recovery must not improve an earlier group and a later unrelated mistake must not degrade it.

Use exactly this score scale:
1 = failure: the action causes or clearly constitutes a major task failure, unsafe event, destructive mistake, or loss of task viability.
2 = suboptimal: the action is noticeably hesitant, inefficient, imprecise, or locally incorrect, but the task remains recoverable.
3 = normal: the action is appropriate, safe, and reasonably efficient for the visible situation.

Return the worst score before the temporal sequence. Worst must equal the minimum of all 32 quality digits. Quality must contain exactly 32 characters from 1, 2, or 3 with no spaces; character i is the score for group i.
For every contiguous group range scored 1 or 2, add one short rationale grounded only in visible evidence. Groups scored 3 get no rationale. Rationale ranges must exactly match the contiguous non-normal runs in Quality.

Return exactly:
Worst: <1, 2, or 3>
Quality:
<32 digits with no spaces>
Rationales:
<first group>-<last group> <short visible-evidence rationale>
<one line per contiguous group range scored 1 or 2, in time order>
\end{rewardprompt}
The user prompt states the task and the clip's absolute time span. A
training example and its target are shown below:
\begin{rewardprompt}{Dense reward: user prompt and target}
Task: assemble the duplofish

Synchronized multiview clip:
<video>

The clip contains 64 sampled frame positions at 4 FPS (16 seconds nominal span) and 32 visible temporal groups of 2 frames each. The sampled positions run from t=353.000 to t=368.733 seconds of the episode. Episode boundaries repeat the nearest available frame.
Predict Worst, all 32 group scores, and rationales for non-normal runs.
---------------------------------------------------------------------------
Worst: 1
Quality:
11111111111133333322222222222233
Rationales:
1-12 fail to place red lego block on the left edge of the lego toy vehicle due to misalignment
19-30 the lego toy vehicle slides while grasping
\end{rewardprompt}
An all-normal window has the target \texttt{Worst:~3}, a string of 32
threes, and an empty \texttt{Rationales:} section.

\textbf{Inference and stitching.}
Decoding is greedy and constrained by a regular expression for the answer
grammar, with at most 640 new tokens. A parsed answer whose \texttt{Worst}
value disagrees with its \texttt{Quality} string is rejected, and an
unparseable answer is replaced by an all-normal window and flagged. Each non-final window retains
groups 1--24 and discards groups 25--32. The final window retains all valid
groups. Let $\mathcal{C}$ be the retained valid groups $(k,g)$, each
contributing the source times $\tau_{k,g}$ of its two frames. A frame at
time $t$ takes the label
\begin{equation}
    a_t=a_{k^\star,g^\star},\qquad
    (k^\star,g^\star)=\operatorname*{lexmin}_{(k,g)\in\mathcal{C}}
    \Big(\min_{\tau\in\tau_{k,g}}\lvert t-\tau\rvert,\;
    -\min(g-1,\,32-g),\;k,\;g\Big),
    \label{eq:dense-stitching}
\end{equation}
that is, the nearest retained group, then the group farthest from its
window's edges, then the earlier window. Consecutive frames with equal
labels form segments, and each non-normal segment reports its most frequent
rationale. Tail trimming keeps at least 4\,s of later video for every
retained group except near the episode end. The reward is retrospective and
intended for offline labeling, analysis, and data filtering.

\subsection{Outcome Reward}
\label{sec:reward-outcome-details}

\textbf{Supervision.}
Outcome supervision combines public, licensed, and internally collected robot recordings. Native five-point labels are retained. Binary success and failure map to 5 and 1. Partial-success values $p\in[0,1]$ are mapped to $s=1+\operatorname{round}(4p)$. Failure categories are mapped only when their outcomes are unambiguous. Training pairs scores with short evidence-grounded rationales when such explanations are available.

\textbf{Frame sampling.}
For an episode with $N$ frames at frame rate $f$, we sample
$m=\min(64,\max(2,\lfloor 10N/f\rfloor))$ frames, rounded down to an even
number, at indices $\operatorname{round}(i(N-1)/(m-1))$ for
$i=0,\dots,m-1$, so the first and last frames are always included.

\textbf{Prompts.}
The two system prompts differ only in their last lines:
\begin{rewardprompt}{Outcome reward: system prompt}
You are a terminal-success reward model for robot trajectories.
Given the task instruction and the complete trajectory video, predict a score from 1 to 5.
A complete failure is 1. A complete success is 5.
Judge only what is visible in the trajectory and follow the requested output format exactly.
[with rationale]
Provide one short, compact rationale followed by the score. Return exactly:
Rationale: <rationale>

Score: <1-5>
[score only]
Do not provide a rationale. Return exactly:
Score: <1-5>
\end{rewardprompt}
\begin{rewardprompt}{Outcome reward: user prompt and target (rationale mode)}
Task: Slide the pink block a short distance toward the robot on the pegboard.

Complete trajectory video:
<video>

Evaluate the terminal success of the task.
---------------------------------------------------------------------------
Rationale: The robot contacted the pink block and nudged it slightly toward the robot base, but the displacement was small and insufficient to complete the requested movement.

Score: 2
\end{rewardprompt}

\textbf{Score distribution.}
Decoding is greedy and constrained to the answer format, with at most 12
new tokens in score-only mode and 160 with a rationale. At the generated
score token, we take the 20 most likely alternatives, keep those that
decode to a digit from 1 to 5, and renormalize their probabilities to
obtain $p_k$ in Eq.~\eqref{eq:outcome-expectation}. RoboRewardBench uses
the decoded score directly. Trajectory ranking uses $\mathbb{E}[s]$.

\subsection{Rubric-Derived Reward}
\label{sec:reward-rubric-details}

\textbf{Rubric format.}
A rubric lists task steps in order. Each step has an action description, a
maximum score, and two to five labeled score levels. Human grades always
take one of these levels, and the maxima sum to one.
Table~\ref{tab:rubric-example} shows an example. Rubrics for internal policy-evaluation episodes contain 2--10 steps, and
rubrics for external episodes 1--4 steps.

\textbf{Supervision.}
Rubrics describe task steps and allowed levels of partial completion. Human grades define the target scores. Training examples pair each video and rubric with a reasoning and answer sequence. The RubricBench evaluation episodes and task names are held out from this pool.

\textbf{Prompt.}
The system prompt is a generic assistant prompt. The user turn lists the
videos, followed by the template below for internal policy-evaluation
episodes. For external sources, the
first paragraph describes the cameras actually present.
\begin{rewardprompt}{Rubric reward: user prompt template (internal policy evaluation)}
Video 1: <video>Video 2: <video>Video 3: <video>
You are an expert evaluator of robot manipulation. You are shown three time-synchronized videos of a bimanual robot performing one task: Video 1 is a top (scene) camera, Video 2 a left-wrist camera, and Video 3 a right-wrist camera.

Score how well the robot performed, using the rubric below. The rubric lists the task's steps; each step has its own maximum number of points and an explicit list of the point values you may award for it. For every step, award exactly one of the values listed for that step -- not a value in between.
Task: {task name}
Instruction to the robot: "{instruction}"

Scoring rubric -- award one of the listed values for each step:
  {i}. {action} (max {maximum})
       award one of: {level} = {points} | ...

Respond with your scores as a JSON array of numbers -- one score per step, in the exact order listed above (a length-{n} array). Each number must be one of the values allowed for that step. Output the array on a single line prefixed with 'ANSWER:' and nothing after it, e.g.
ANSWER: {list of the step maxima}
\end{rewardprompt}
The answer is parsed from the last \texttt{ANSWER:} line, and each value is
clipped to $[0,\max_j]$.

\subsection{Co-Training}
\label{sec:reward-training-details}

\textbf{Shared supervised adaptation.}
The released reward model uses SFT only. Outcome, dense, and rubric examples are shuffled together without task-family reweighting. Rubric answers include reasoning, while outcome and dense answers use direct structured outputs. A shared Qwen3.5-27B model updates the language model, visual merger, and visual encoder, using a lower learning rate for the visual encoder. Training minimizes output-token cross-entropy (Eq.~\eqref{eq:annotation-training}). The input budget accommodates the longer rubric answers.

\subsection{Interface Examples}
\begin{figure}[H]
\centering
\includegraphics[width=0.93\linewidth]{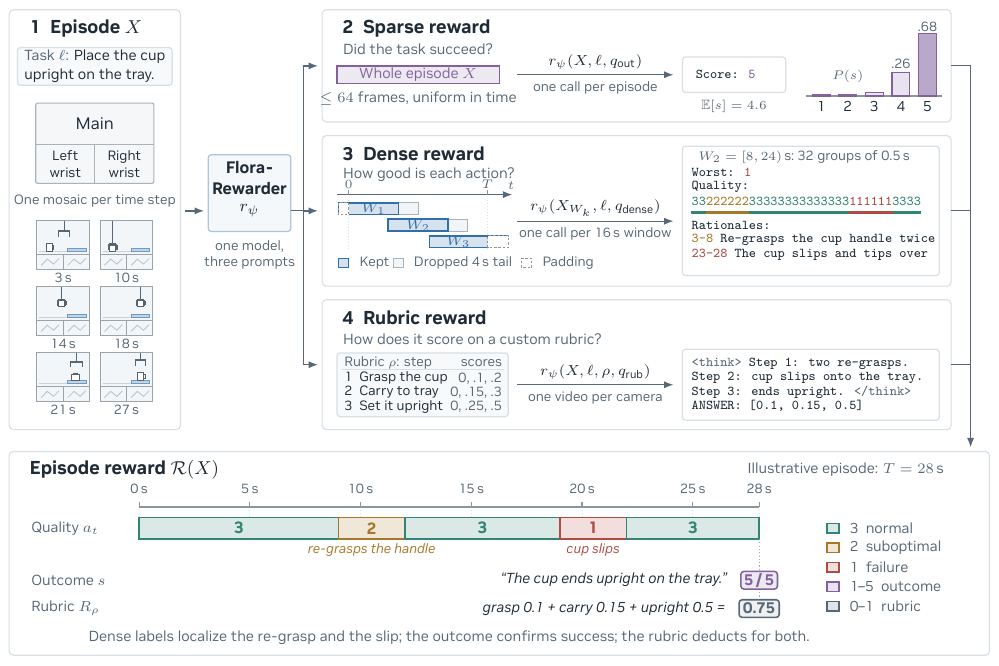}
\caption{\textbf{Three interfaces, one reward model.} Outcome prompts score episode completion, dense prompts label temporal groups with rationales, and rubric prompts assign discrete credit to user-defined task steps.}
\label{fig:reward-model}
\end{figure}

\phantomsection
\label{app:rubrics}

\begin{table}[H]
\centering\floraTableStyle
\renewcommand{\arraystretch}{1.1}
\begin{tabularx}{\linewidth}{@{}p{0.29\linewidth} c X@{}}
\toprule
\textbf{Step} & \textbf{Score} & \textbf{Criterion} \\
\midrule
\multirow{3}{=}{\textbf{1.} Firmly grasp apple}
  & 0.0 & Did not grasp the apple \\
  & 0.2 & Touched but did not grab the apple \\
  & 0.4 & Firmly grasped the apple \\
\midrule
\multirow{3}{=}{\textbf{2.} Pick up knife}
  & 0.0 & Did not pick up the knife \\
  & 0.15 & Grasped the knife but dropped it \\
  & 0.3 & Picked up the knife with a secure grip \\
\midrule
\multirow{3}{=}{\textbf{3.} Cut apple in half}
  & 0.0 & Did not cut the apple \\
  & 0.15 & Partially cut the apple \\
  & 0.3 & Cut the apple fully in half \\
\midrule
\textbf{Maximum total} & \textbf{1.0} & \\
\bottomrule
\end{tabularx}
\caption{Sample scoring rubric for an apple-cutting task.}
\label{tab:rubric-example}
\end{table}

\clearpage
\section{Related Work}
\label{sec:related}

\textbf{Captioning for robotics.} Commercial services such as Scale AI's Physical AI Data Engine~\cite{scale2025physicalai} provide data collection and model-assisted annotation for robotics. Open-source alternatives have recently emerged. Macrodata Labs' Refiner~\cite{macrodata2026refiner} provides subtask segmentation and reward scoring along with the WGO-Bench benchmark~\cite{wgobench2026}. LeRobot~\cite{cadene2024lerobot,lerobot2026v060} uses an off-the-shelf VLM to write subtasks, plans, and VQA pairs into datasets, and uses pretrained reward models such as Robometer~\cite{robometer2026} to label per-frame progress. Their captioning pipelines compose existing VLMs with task-specific prompts and processing. General video-language models such as Pegasus~\cite{twelvelabs2024pegasus} support video understanding and summarization, while DIAL and ECoT generate language supervision for robot policy learning~\cite{xiao2022dial,zawalski2024ecot}.

\textbf{Open-weight VLMs as annotators.} Open-weight VLMs provide backbones for locally deployed annotators. General-purpose candidates include Gemma~\cite{gemma2025gemma3,gemma2026gemma4}, Qwen3-VL~\cite{bai2025qwen3vl}, InternVL3.5~\cite{wang2025internvl35}, GLM-4.5V~\cite{glm2025glm45v}, and Kimi~\cite{kimi2025kimivl,kimi2026k25}. CapRL~\cite{xing2025caprl,caprlpp2026} shows that captioners can be optimized with RL for downstream utility. \name builds on these advances but targets a narrower goal: annotations of robot and human data that are accurate, consistent, and cheap enough to cover large pretraining corpora.

\textbf{Reward models for robotics.} Robot reward modeling has progressed from visual representations~\cite{ma2022vip,ma2023liv} and zero-shot VLM similarity~\cite{rocamonde2023vlmrm,sontakke2023roboclip}, to in-context progress estimation~\cite{ma2024gvl}, and most recently to fine-tuned reward models. RoboReward~\cite{lee2026roboreward} synthesizes failures to train episode-level rewards, Robometer~\cite{robometer2026} combines frame-level progress with trajectory preferences, and Robo-Dopamine~\cite{tan2025robodopamine} targets process rewards for high-precision manipulation.

\end{document}